\documentclass{article} 
\usepackage{iclr2022_conference,times}
\usepackage[utf8]{inputenc} 
\usepackage[T1]{fontenc}    
\usepackage{hyperref}       
\usepackage{url}            
\usepackage{booktabs}       
\usepackage{amsfonts}       
\usepackage{nicefrac}       
\usepackage{microtype}      
\usepackage{colortbl}
\usepackage{graphicx}
\usepackage{subfigure} 
\usepackage{pifont}

\usepackage{amsmath,amsfonts,bm}

\def\eqref#1{(\ref{#1})}

\def\1{\bm{1}}

\DeclareMathAlphabet{\mathsfit}{\encodingdefault}{\sfdefault}{m}{sl}
\SetMathAlphabet{\mathsfit}{bold}{\encodingdefault}{\sfdefault}{bx}{n}

\usepackage{adjustbox}
\usepackage{lipsum}
\usepackage{color}
\usepackage{wrapfig}
\usepackage{booktabs}
\usepackage{multirow,mathtools} 
\usepackage{algorithm,algpseudocode}
\usepackage{threeparttable}

 \algnewcommand{\algorithmicforeach}{\textbf{for each}}
\algdef{SE}[FOR]{ForEach}{EndForEach}[1]
  {\algorithmicforeach\ #1\ \algorithmicdo}
  {\algorithmicend\ \algorithmicforeach}
  \usepackage{xcolor}
\definecolor{lightblue}{rgb}{0.68, 0.85, 0.9}
\definecolor{lightgreen}{rgb}{0.56, 0.93, 0.56}
\definecolor{lightskyblue}{rgb}{0.53, 0.81, 0.98}
\definecolor{non-photoblue}{rgb}{0.64, 0.87, 0.93}
\definecolor{magicmint}{rgb}{0.67, 0.94, 0.82}
\definecolor{mossgreen}{rgb}{0.68, 0.87, 0.68}
\definecolor{salmon}{rgb}{1.0, 0.55, 0.41}
\definecolor{babypink}{rgb}{0.96, 0.76, 0.76}

\DeclareMathOperator*{\minimize}{\text{minimize}}

\DeclareMathAlphabet\mathbfcal{OMS}{cmsy}{b}{n}
\newcommand{\Def}[0]{\mathrel{\mathop:}=}

\newcommand{\btheta}{\boldsymbol{\theta}}

\newcommand{\bdelta}{\boldsymbol{\delta}}

\newcommand{\mycomment}[1]{}
\newcommand{\IA}{\text{PA}}
\newcommand{\DeRED}{\text{CDD-RED}}

\newcommand{\SubSection}[1]{\vspace{-2mm} \subsection{#1} \vspace{-2mm}}

\newcommand{\AdvD}{\text{AD}}

\newcommand{\AtrA}{\text{IAA}}

\title{
Reverse Engineering of Imperceptible   Adversarial Image Perturbations}

\author{%
  Yifan Gong$^1$\thanks{Equal contributions.}, Yuguang Yao$^{2*}$, Yize Li$^{1}$, Yimeng Zhang$^{2}$, Xiaoming Liu$^{2}$, Xue Lin$^{1}$, Sijia Liu$^{2,3}$
 \\
  $^{1}$ Northeastern University, $^{2}$ Michigan State University, $^{3}$ MIT-IBM Watson AI Lab, IBM Research
\\
  \texttt{\{gong.yifa, li.yize, xue.lin\}@northeastern.edu}, \\
  \texttt{\{yaoyugua, zhan1853, liusiji5\}@msu.edu, liuxm@cse.msu.edu}}

\iclrfinalcopy 
\begin{document}

\maketitle

\begin{abstract}
It has been well recognized that neural network based image classifiers are easily fooled by images with tiny perturbations crafted by an adversary. There has been a vast volume of research to generate and defend such adversarial attacks. However, the following problem is left unexplored: \textit{How to reverse-engineer adversarial perturbations from an adversarial image?}
This leads to a new adversarial learning paradigm---Reverse Engineering of Deceptions (RED). 
If successful, RED allows us to estimate adversarial perturbations and recover the original images. 
However, carefully crafted, tiny adversarial perturbations are difficult to recover by optimizing a unilateral RED objective. For example, the pure image denoising method may overfit to minimizing the reconstruction error but hardly preserves the classification properties of the true adversarial perturbations. 
To tackle this challenge, we formalize the RED problem and identify a set of principles crucial to the RED approach design. 
Particularly, we find that prediction alignment and proper data augmentation (in terms of spatial transformations) are two criteria to achieve a generalizable  RED approach. 
By integrating these RED principles with image denoising, we propose a new \underline{C}lass-\underline{D}iscriminative \underline{D}enoising based \underline{RED} framework, termed {{\DeRED}}.
Extensive experiments demonstrate the effectiveness of {\DeRED} under different evaluation metrics (ranging from the pixel-level,   prediction-level to the attribution-level alignment) and a variety of attack generation methods ({\it e.g.}, FGSM, PGD, CW,  AutoAttack, and adaptive attacks). Codes are available at \href{https://github.com/Yifanfanfanfan/Reverse-Engineering-of-Imperceptible-Adversarial-Image-Perturbations}{\texttt{link}}. 
%
\end{abstract}

\vspace{-3mm}
\section{Introduction}


Deep neural networks (DNNs) are susceptible
to adversarially-crafted tiny input perturbations during inference.
Such imperceptible   perturbations, {\it a.k.a.}~adversarial attacks, could cause  DNNs to draw manifestly wrong conclusions. 
The existence of adversarial attacks was first uncovered in the domain of image classification \citep{Goodfellow2015explaining,carlini2017towards,papernot2016limitations}, and was then rapidly extended to  the other domains, such as object detection \citep{xie2017adversarial,serban2020adversarial}, language modeling \citep{cheng2020seq2sick,srikant2021generating}, and medical machine learning \citep{finlayson2019adversarial,antun2020instabilities}. Despite different applications, the underlying attack  formulations and generation methods commonly obey the ones used in image classification. 

A vast volume of existing works have been devoted to designing defenses against such attacks, mostly focusing on either detecting adversarial examples \citep{grosse2017statistical,yang2020ml,metzen2017detecting,meng2017magnet,wojcik2020adversarial} or acquiring  adversarially robust DNNs \citep{madry2017towards,zhang2019theoretically,wong2017provable,salman2020denoised,wong2020fast,carmon2019unlabeled,shafahi2019adversarial}.
Despite the plethora of prior work on adversarial defenses, it seems impossible to achieve `perfect'    robustness. 
Given the fact that adversarial attacks are inevitable~\citep{shafahi_are_2020}, we ask whether  or not an adversarial  attack can be  reverse-engineered so that one can estimate the  adversary's  information ({\it e.g.}, adversarial perturbations) behind the attack instances.
The above problem is referred to as \textit{Reverse Engineering of Deceptions (RED)}, fostering  a new adversarial learning regime. 
The development of RED technologies will also
enable the adversarial situation awareness in high-stake applications.

To the best of our knowledge, few work studied the RED problem. The most relevant one  that we are aware of is \citep{pang_advmind_2020}, which proposed the so-called query of interest (QOI) estimation model to  
 infer the   adversary’s   target class by model queries.
 However, the work \citep{pang_advmind_2020} was restricted to the black-box attack scenario and thus lacks a general formulation of RED. Furthermore, it has  not    built
  a complete RED pipeline, which  should not only provide a solution to estimating the adversarial example but also  formalizing  evaluation metrics to comprehensively measure the  performance of RED.
    In this paper, we aim to take a solid step towards addressing the RED problem.

\SubSection{Contributions}
The main contributions of our work is listed below. 

\noindent $\bullet$ 
We formulate the Reverse Engineering of Deceptions (RED) problem that is able to estimate adversarial perturbations and 
provides the feasibility of inferring the intention of an adversary, {\it e.g.}, `adversary saliency regions' of an adversarial image.

\noindent $\bullet$ 
We identify a series of RED principles to  
effectively estimate the adversarially-crafted tiny perturbations.
We find that the class-discriminative ability is crucial to evaluate the RED performance.
We also find that 
data augmentation, \textit{e.g.}, spatial transformations, is another key to  improve the RED result.
 Furthermore, we integrate the developed RED principles into 
image denoising  and 
propose a denoiser-assisted RED approach.


\noindent $\bullet$ 
We  
build a comprehensive evaluation pipeline to quantify the RED performance from different perspectives, such as pixel-level reconstruction error, prediction-level alignment, and attribution-level adversary saliency region recovery. 
With an extensive experimental study,
 we show that, 
 compared to image denoising baselines,
 our proposal
 yields a consistent   improvement across diverse RED evaluation metrics and attack generation methods, {\it e.g.},  FGSM \citep{Goodfellow2015explaining}, CW \citep{carlini2017towards}, PGD \citep{madry2017towards} and AutoAttack \citep{croce2020reliable}.

\SubSection{Related work}

\paragraph{Adversarial attacks.}
Different types of adversarial attacks have been proposed, ranging from digital attacks   \citep{Goodfellow2015explaining,carlini2017towards,madry2017towards,croce2020reliable,xu2019structured,chen2017ead,xiao2018spatially}
 to physical attacks \citep{eykholt2018robust,li2019adversarial,athalye18b,chen2018shapeshifter,xu2019evading}. The former  
gives the most fundamental threat model that
commonly deceives DNN models during inference by crafting imperceptible adversarial perturbations.
The latter extends the former to fool the victim models in the physical environment. Compared to digital attacks, physical attacks require  much larger perturbation strengths to enhance the adversary's resilience to various physical conditions such as lightness and object   deformation \citep{athalye18b,xu2019evading}.

In this paper, we focus on $\ell_p$-norm ball constrained   attacks, {\it a.k.a.}~$\ell_p$ attacks, for $p \in \{ 1,2,\infty\}$, most widely-used in digital attacks. 
Examples  include FGSM \citep{Goodfellow2015explaining}, PGD \citep{madry2017towards}, CW \citep{carlini2017towards}, and the recently-released attack benchmark AutoAttack \citep{croce2020reliable}. 
Based on the adversary's intent, $\ell_p$ attacks are further divided into untargeted attacks and targeted attacks, where in contrast to the former, the latter designates the (incorrect) prediction label of a victim model.
When an adversary has no access to victim models' detailed information (such as architectures and model weights), $\ell_p$ attacks can   be further generalized to   black-box attacks  by leveraging either surrogate victim models \citep{papernot_practical_2017,papernot_transferability_2016,dong_evading_2019,liu_delving_2017} or  input-output queries from   the original black-box models \citep{chen2017zoo,liu2018signsgd,cheng2019sign}. 

\paragraph{Adversarial defenses.}
To improve the robustness of DNNs, a variety of approaches have been proposed
to defend against   $\ell_p$ attacks.
One line of research focuses on enhancing the robustness of DNNs during training, {\it e.g.},
adversarial training \citep{madry2017towards}, TRADES \citep{zhang2019theoretically}, randomized smoothing \citep{wong2017provable}, and their variants \citep{salman2020denoised,wong2020fast,carmon2019unlabeled,shafahi2019adversarial,uesato2019labels,chenliu2020cvpr}.
Another line of research is to detect adversarial attacks without altering the victim model or the training process. 
The key  technique is to differentiate between benign and adversarial examples by measuring their `distance.'
Such a distance measure has been defined in the input space via pixel-level reconstruction error~\citep{meng2017magnet,liao_defense_2018}, in the intermediate layers via neuron activation anomalies \citep{xu2019interpreting}, and in the  logit space by tracking the sensitivity of deep feature attributions to input perturbations \citep{yang2020ml}. 

In contrast to RED, \textit{adversarial detection is a relatively simpler problem} as a roughly approximated distance possesses detection-ability \citep{meng2017magnet,luo2015foveation}. 
  Among the existing adversarial defense techniques, the recently-proposed Denoised Smoothing (DS) method \citep{salman2020denoised} is more related to ours.  In \citep{salman2020denoised}, an image denoising network   is  prepended to  an existing victim model so that the augmented   system can be performed as a smoothed image classifier with certified robustness.  
  Although DS is not designed for RED, its denoised output can be regarded as a benign example estimate.
  The promotion of classification stability in DS also motivates us to design the RED methods with class-discriminative ability. 
  Thus, DS will be  a main baseline approach for comparison.   Similar to our RED setting, the concurrent work \citep{souri2021identification} also identified the feasibility of estimating adversarial perturbations from adversarial examples.
  

  
  
  
  

\vspace{-3mm}
\section{Reverse Engineering of Deceptions: Formulation and Challenges} \label{sec:red_formulation_challenges}

\vspace{-2mm}
In this section, we first introduce the threat model of our interest: adversarial attacks on images. 
Based on that, we formalize the Reverse Engineering of Deceptions (RED) problem and demonstrate its challenges through some `warm-up' examples.
\vspace{-2mm}
\paragraph{Preliminaries on threat model.}
We focus on $\ell_p$ attacks, where the \textit{adversary's goal} is to generate imperceptible input perturbations to fool a well-trained image classifier.
Formally, let $\mathbf x$ denote a benign image, and $\boldsymbol \delta$ an additive   perturbation variable. Given a victim classifier $f$ and a perturbation strength tolerance $\epsilon$ (in terms of, {\it e.g.}, $\ell_\infty$-norm constraint  $ \| \bdelta \|_\infty  \leq   
\epsilon$), the desired \textit{attack generation algorithm} $\mathcal A$    then seeks the optimal   $\boldsymbol \delta$ subject to the perturbation constraints. Such an attack generation process is   denoted by 
$\boldsymbol \delta = \mathcal A (\mathbf x, f, \epsilon)$, resulting in an adversarial example   $\mathbf x^{\prime} = \mathbf x + \boldsymbol \delta$.
{Here $\mathcal A$ can be fulfilled by different attack methods, e.g., FGSM~\citep{Goodfellow2015explaining}, CW~\citep{carlini2017towards}, PGD~\citep{madry2017towards}, and AutoAttack \citep{croce2020reliable}.}
\vspace{-2mm}
\paragraph{Problem formulation of RED.}
Different from conventional defenses to detect or reject   adversarial instances \citep{pang_advmind_2020,liao_defense_2018,shafahi_are_2020,niu_limitations_2020}, RED aims to address the following question.
\begin{center}
	\vspace{-2mm}
	\setlength\fboxrule{0.0pt}
	\noindent\fcolorbox{black}[rgb]{0.95,0.95,0.95}{\begin{minipage}{0.98\columnwidth}
				\vspace{-0.07cm}
				{\bf (RED problem)}  Given an adversarial instance, can we reverse-engineer the adversarial perturbations $\boldsymbol \delta$, and infer    the adversary's objective and knowledge, {\it e.g.}, true image class behind deception and adversary saliency image region? 
				\vspace{-0.07cm}
	\end{minipage}}
	\vspace{-2mm}
\end{center}

Formally,
we aim to recover  $\boldsymbol \delta$  from an adversarial example $\mathbf x^{\prime}$
under the prior knowledge   
of the  victim model $f$ or its substitute  $\hat{f}$ if the former is a black box.
We denote the RED operation as $\boldsymbol \delta = \mathcal R(\mathbf x^{\prime}, \hat {f})$, 
which covers the white-box   scenario ($\hat {f} = f$) as a special case. 
We   propose to learn a
parametric model $\mathcal D_{\boldsymbol \theta}$ ({\it e.g.}, a denoising  neural network that we will focus on)
as an approximation of $\mathcal R$ through
  a  training
dataset of adversary-benignity pairs $ \Omega = \{ (\mathbf x^{\prime}, \mathbf x) \}$. 
 Through $\mathcal D_{\boldsymbol \theta}$, 
RED will provide   a \textbf{benign example estimate} $\mathbf x_{\mathrm{RED}}$ and a   \textbf{adversarial example estimate} $\mathbf x_{\mathrm{RED}}^\prime$ as below:

{
\vspace*{-6mm}
\small{
\begin{align}\label{eq: RED_results}
    \mathbf x_{\mathrm{RED}} = \mathcal D_{\btheta} (\mathbf x^{\prime}), \quad \mathbf x_{\mathrm{RED}}^\prime = \underbrace{ \mathbf x^{\prime} -\mathbf x_{\mathrm{RED}} }_\text{perturbation estimate} + \mathbf x,
\end{align}
}}%

where a   \textbf{perturbation estimate} is given by subtracting the RED's output with its input, 
$ \mathbf x^{\prime} - \mathcal D_{\boldsymbol \theta}(\mathbf x^{\prime})$.


\begin{wrapfigure}{r}{60mm}
\vspace{-5mm}
\centering{
\begin{tabular}{c}
\hspace*{-3mm}
\includegraphics[width=0.4\textwidth]{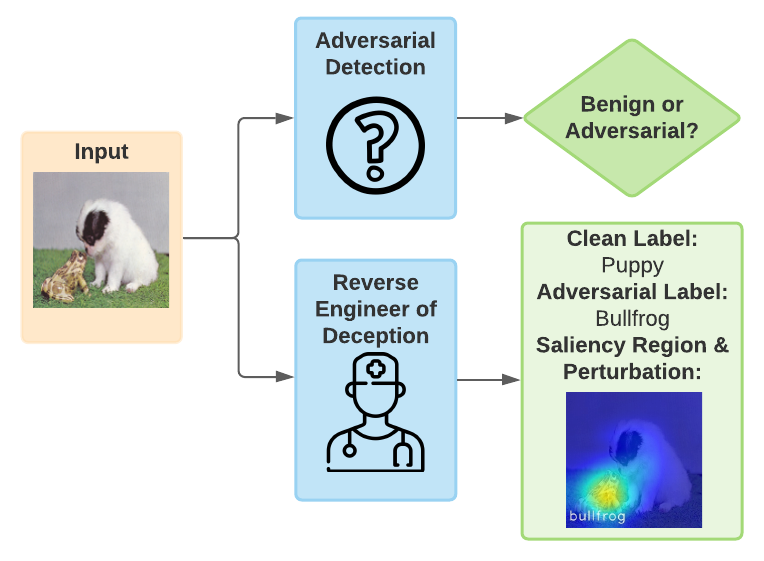}
\end{tabular}
}
\vspace{-5mm}
\caption{\footnotesize{Overview of {RED} versus {\AdvD}.
}}
\vspace{-4mm}
\label{fig: REDvsAD}
\end{wrapfigure}
We \textbf{highlight} that RED yields a new  defensive approach aiming to `diagnose' the perturbation details of an existing adversarial example in a post-hoc, forensic manner. 
This is different from   {adversarial detection ({\AdvD})}. Fig.\ref{fig: REDvsAD} provides a visual comparison of {RED} with {\AdvD}. 
 Although {\AdvD} is  also   designed in a post-hoc manner, it  aims  to determine whether  an input is an adversarial example for a victim model based on certain statistics on  model features or logits. Besides, 
{\AdvD} might    be used as a pre-processing step of {RED}, where the former provides `detected' adversarial examples for  fine-level RED diagnosis.
{In our experiments, we will also show that the outputs of RED can be   leveraged to guide the design of adversarial detection. In this sense, {RED} and {\AdvD}  are  complementary building blocks within a closed loop.}

\paragraph{Challenges of RED}
In this work, we  will specify the   RED model $\mathcal D_{\btheta}$  as a {denoising network}. However,
it is highly non-trivial to design a proper  denoiser for RED. 
Speaking at a high level, 
there exist two {main challenges}. 
First, unlike 
the conventional image denoising strategies~\citep{DNCNN_denoiser},
the design of an RED-aware denoiser needs to take into account the effects of    victim models    and   data properties of  adversary-benignity pairs. 
Second,  it might be insufficient to merely minimize the reconstruction error as the adversarial perturbation is  
finely-crafted~\citep{niu_limitations_2020}. Therefore, either under- or over-denoising   will lead to poor RED performance.







\section{{RED Evaluation Metrics and   Denoising-Only Baseline}} \label{sec:evaluation_metric}

Since RED  is different from existing defensive approaches, we first develop new performance metrics of RED, ranging from
pixel-level reconstruction error to attribution-level adversary saliency region. 
We next leverage the proposed performance metrics to  demonstrate why a pure image denoiser is incapable of fulfilling RED. 

\paragraph{RED evaluation metrics.}
   Given a learned RED model $\mathcal D_{\btheta}$, the RED performance will be evaluated over a testing dataset $(\mathbf x^{\prime}, \mathbf x)  \in \mathcal D_{\mathrm{test}}$; see implementation details in Sec.\,\ref{sec: exp}. Here, $\mathbf x^{\prime}$ is used as the testing  input of the RED model, 
   and $\mathbf x$ is the associated ground-truth benign example for comparison.  The benign example estimate $\mathbf x_{\mathrm{RED}}$ and adversarial example estimate  $\mathbf x_{\mathrm{RED}}^\prime$ are obtained following \eqref{eq: RED_results}. 
   RED evaluation pipeline is conducted from  the following aspects: \ding{172} pixel-level reconstruction error, \ding{173} prediction-level inference  alignment, and \ding{174} attribution-level adversary saliency region.

\noindent \ding{226}    \ding{172} \textbf{Pixel-level}: Reconstruction error given by
 $d(\mathbf x, \mathbf x_\mathrm{RED})=
\mathbb E_{(\mathbf x^\prime, \mathbf x) \in \mathcal D_{\mathrm{test}}} [\| \mathbf x_{\mathrm{RED}} - \mathbf x \|_2 ]$.   



\noindent \ding{226} \ding{173} \textbf{Prediction-level}: 
Prediction alignment (\IA) between the pair of \textit{benign} example and  its estimate $(\mathbf x_{\mathrm{RED}}, \mathbf x)$ and {\IA} between the pair of \textit{adversarial} example and its estimate $(\mathbf x_{\mathrm{RED}}^\prime, \mathbf x^\prime)$, given by

{
\small{
\vspace*{-5mm}
\begin{align}
\text{\IA}_{\mathrm{benign}} =  \frac{ \mathrm{card}( \{ (\mathbf x_{\mathrm{RED}}, \mathbf x) \, | \,F(\mathbf x_{\mathrm{RED} } ) = F(\mathbf x  ) \} ) }{\mathrm{card}(\mathcal D_{\mathrm{test}})},~
\text{\IA}_{\mathrm{adv}}=  \frac{ \mathrm{card}( \{ (\mathbf x_{\mathrm{RED}}^\prime, \mathbf x^\prime) \, | \, F(\mathbf x_{\mathrm{RED} }^\prime ) = F(\mathbf x^\prime  ) \} ) }{\mathrm{card}(\mathcal D_{\mathrm{test}})} \nonumber 
\end{align}
}}%
where $\mathrm{card}(\cdot)$ denotes a cardinality function of a set and $F$ refers to the prediction label provided by the victim model $f$. 

\noindent \ding{226}    \ding{174} \textbf{Attribution-level}:
Input attribution alignment ({\AtrA}) between  the benign pair $(\mathbf x_{\mathrm{RED}}, \mathbf x)$ and   between the adversarial pair   $(\mathbf x_{\mathrm{RED}}^\prime, \mathbf x^\prime)$. In this work, we adopt 
GradCAM \citep{selvaraju_grad-cam_2020} to attribute the predictions of classes back to input 
saliency regions.
The rationale behind {\AtrA} is that the unnoticeable adversarial perturbations (in the pixel space) can introduce  an evident input attribution   discrepancy with respect to (w.r.t.) the true label $y$ and the adversary's target label $y^\prime$ \citep{boopathy2020proper,xu2019structured}. Thus, an accurate RED should be able to
erase the adversarial attribution effect through $\mathbf x_{\mathrm{RED}}$, and   
estimate the adversarial intent through the saliency region of $\mathbf x_{\mathrm{RED}}^\prime$ (see Fig.\,\ref{fig: REDvsAD} for   illustration).

\paragraph{Denoising-Only (DO) baseline.}
\begin{wrapfigure}{r}{70mm}
  \vspace{-10mm}
  \centering
  \begin{adjustbox}{max width=0.5\textwidth }
  \begin{tabular}{@{\hskip 0.00in}c @{\hskip 0.00in}c | @{\hskip 0.00in} @{\hskip 0.02in} c @{\hskip 0.02in} | @{\hskip 0.02in} c @{\hskip 0.02in} |@{\hskip 0.02in} c @{\hskip 0.02in} |@{\hskip 0.02in} c @{\hskip 0.02in} 
  }
& 
\colorbox{lightgray}{ \textbf{Input image}}
&
\colorbox{lightgray}{ \textbf{DO}} 
&
\colorbox{lightgray}{  \textbf{Groundtruth}}

\\
 \begin{tabular}{@{}c@{}}  

\rotatebox{90}{\parbox{10em}{\centering  \textbf{\large{Benign example $\mathbf x$/$\mathbf x_{\mathrm{RED}}$}}}}
 \\

\rotatebox{90}{\parbox{10em}{\centering  \textbf{\large{Adv. example $\mathbf x^\prime$/$\mathbf x_{\mathrm{RED}}^\prime$} 
}}}

\end{tabular} 
& 
\begin{tabular}{@{\hskip 0.02in}c@{\hskip 0.02in}}
\\
 \begin{tabular}{@{\hskip 0.02in}c@{\hskip 0.02in} }
 \parbox[c]{10em}{\includegraphics[width=10em]{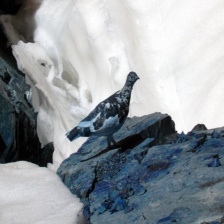}

 }  
\end{tabular}
 \\
 \begin{tabular}{@{\hskip 0.02in}c@{\hskip 0.02in}}
 \parbox[c]{10em}{\includegraphics[width=10em]{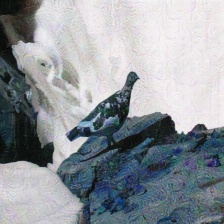}}  
\end{tabular}
\end{tabular}
&
\begin{tabular}{@{\hskip 0.02in}c@{\hskip 0.02in}}
 \begin{tabular}{@{\hskip 0.02in}c@{\hskip 0.02in}c@{\hskip 0.02in} }
      \begin{tabular}{@{\hskip 0.00in}c@{\hskip 0.00in}}
     \parbox{10em}{\centering   $I(\cdot, y)$}  
    \end{tabular} 
     &  
      \begin{tabular}{@{\hskip 0.00in}c@{\hskip 0.00in}}
       \parbox{10em}{\centering   $I(\cdot, y^\prime )$} 
    \end{tabular} 
\end{tabular}\\
 \begin{tabular}{@{\hskip 0.02in}c@{\hskip 0.02in}c@{\hskip 0.02in} }
 \parbox[c]{10em}{\includegraphics[width=10em]{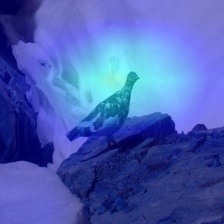}

 } &    \parbox[c]{10em}{\includegraphics[width=10em]{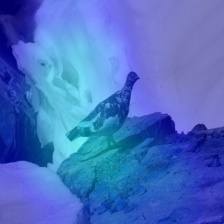}} 
\end{tabular}
 \\
 \begin{tabular}{@{\hskip 0.02in}c@{\hskip 0.02in}c@{\hskip 0.02in} }
 \parbox[c]{10em}{\includegraphics[width=10em]{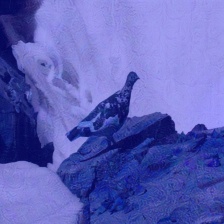}} &    \parbox[c]{10em}{\includegraphics[width=10em]{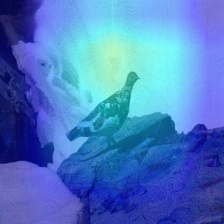}} \\
\end{tabular}
\end{tabular}
&
\begin{tabular}{@{\hskip 0.02in}c@{\hskip 0.02in}}
 \begin{tabular}{@{\hskip 0.02in}c@{\hskip 0.02in}c@{\hskip 0.02in} }
      \begin{tabular}{@{\hskip 0.00in}c@{\hskip 0.00in}}
     \parbox{10em}{\centering   $I(\cdot, y)$}  
    \end{tabular} 
     &  
      \begin{tabular}{@{\hskip 0.00in}c@{\hskip 0.00in}}
       \parbox{10em}{\centering   $I(\cdot, y^\prime )$} 
    \end{tabular} 
\end{tabular}\\
 \begin{tabular}{@{\hskip 0.02in}c@{\hskip 0.02in}c@{\hskip 0.02in} }
 \parbox[c]{10em}{\includegraphics[width=10em]{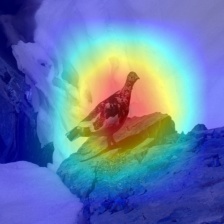}} &    \parbox[c]{10em}{\includegraphics[width=10em]{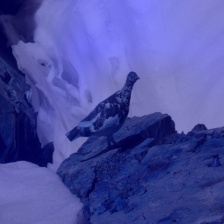}} 
\end{tabular}
 \\
 \begin{tabular}{@{\hskip 0.02in}c@{\hskip 0.02in}c@{\hskip 0.02in} }
 \parbox[c]{10em}{\includegraphics[width=10em]{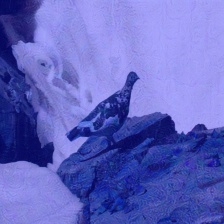}} &    \parbox[c]{10em}{\includegraphics[width=10em]{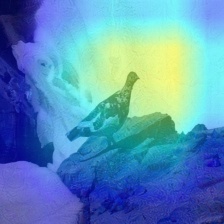}} 
 \\
\end{tabular}
\end{tabular}
\end{tabular}
  \end{adjustbox}
  \vspace*{-3mm}
\caption{\footnotesize{
IAA  of DO compared with ground-truth.
}}
\label{fig: DOIAA}
\end{wrapfigure}
We further show that how a pure image denoiser, a `must-try' baseline, is insufficient of  
tackling the RED problem.
This failure case drive us to rethink the denoising strategy through the lens of RED. 
First, we obtain the denoising network by minimizing the reconstruction error:

{\small
\vspace*{-2.5mm}
\begin{align}
    \begin{array}{ll}
\displaystyle \minimize_{\btheta}         & \ell_{\mathrm{denoise}}(\btheta; \Omega) \Def \mathbb E_{(\mathbf x^{\prime}, \mathbf x) \in \Omega }\|  \mathcal D_{\btheta} (\mathbf x^{\prime}) - \mathbf x \|_1, 
    \end{array}
    \label{eq: MAE_only}
\end{align}
}%
where a Mean Absolute Error (MAE)-type loss is used for denoising \citep{liao_defense_2018}, and the creation of training dataset $\Omega$ is illustrated in Sec.\,\ref{sec: exp_setup}.
Let us then evaluate the performance of {DO} through the non-adversarial prediction alignment $\mathrm{PA}_\mathrm{benign}$ and {\AtrA}. We find that
$\mathrm{PA}_\mathrm{benign} = 42.8\%$ for DO. And
 Fig.\,\ref{fig: DOIAA} shows  the {{\AtrA}} performance   of DO w.r.t. an input example. As we can see,  DO is not capable of exactly recovering the adversarial saliency regions compared to the ground-truth adversarial perturbations.
 These suggest that
 DO-based RED lacks
 the reconstruction ability at the prediction and the attribution levels. {Another naive approach is performing adversarial attack back to $x'$, yet it requires additional assumptions and might not precisely recover the ground-truth perturbations. The detailed limitations are discussed in Appendix \ref{sec:naive_red}.}

\section{{Class-Discriminative Denoising for RED}}
\label{sec: approach}

\begin{wrapfigure}{r}{70mm}
    \vspace*{-5mm}
\centerline{
\includegraphics[width=.46\textwidth,height=!]{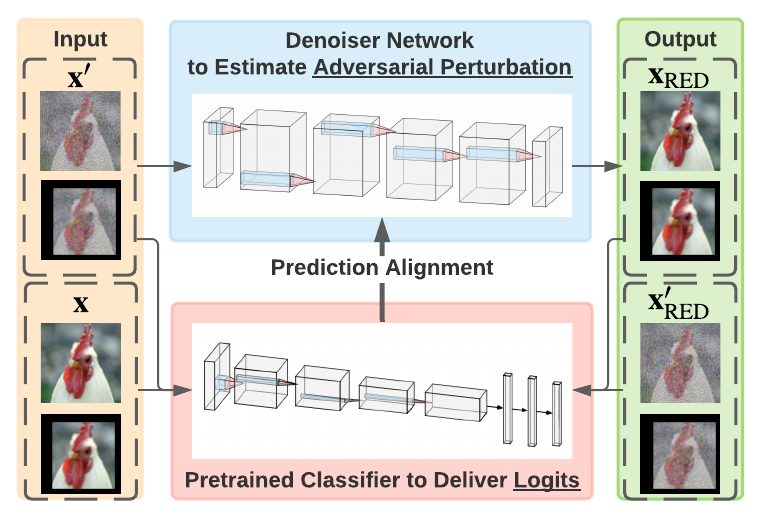}
}
\vspace*{-5mm}
\caption{\footnotesize{
{\DeRED} overview.  
}}
  \label{fig: overview}
  \vspace*{-5mm}
\end{wrapfigure}

In this section, we propose a novel 
\underline{C}lass-\underline{D}iscriminative 
\underline{D}enoising based \underline{RED} approach termed {{\DeRED}}; see Fig.\,\ref{fig: overview} for an overview. 
{{\DeRED}} contains two key  components. First, we propose a  
{\IA}
regularization   to enforce the prediction-level stabilities of both estimated benign example  $\mathbf x_{\mathrm{RED}}$ and adversarial example  $\mathbf x_{\mathrm{RED}}^\prime$ with respect to their true counterparts $\mathbf x$ and $\mathbf x^\prime$, respectively. 
Second, 
we propose a data augmentation strategy
to improve the  {{RED}}'s generalization  without losing  its class-discriminative ability. 


\paragraph{Benign and adversarial prediction alignment.}
To accurately estimate the adversarial perturbation from an adversarial instance, the lessons from the DO approach suggest to preserve the class-discriminative ability of   RED estimates to align with the original predictions, given by $\mathbf x_{\mathrm{RED}}$ vs. $\mathbf x$, and $\mathbf x_{\mathrm{RED}}^\prime$ vs. $\mathbf x^\prime$.
Spurred by that, 
the training objective  
of {\DeRED} is required 
   not only to  minimize the reconstruction error like \eqref{eq: MAE_only} but also to maximize  {{\IA}}, namely,
`clone' the class-discriminative ability of original data. 
To achieve this goal,  we augment the  denoiser  $\mathcal D_{\btheta}$ with a known classifier $\hat{f}$ to generate predictions of 
estimated  benign and adversarial examples (see Fig.\,\ref{fig: overview}), \textit{i.e.}, $\mathbf x_{\mathrm{RED}}$ and $\mathbf x_{\mathrm{RED}}^{\prime}$ defined in \eqref{eq: RED_results}. 
By contrasting $\hat {f} (\mathbf x_{\mathrm{RED}})$ with 
$\hat{f}(\mathbf x) $, and  $\hat {f} (\mathbf x_{\mathrm{RED}}^\prime)$ with 
$\hat{f}(\mathbf x^{\prime}) $, we can promote {\IA} by minimizing  the prediction gap between true examples and estimated ones:

{\vspace*{-3.5mm}
\small\begin{align}\label{eq: IA}
    \ell_{\text{\IA}} ( \btheta; \Omega) =  
   \mathbb E_{(\mathbf x^\prime, \mathbf x) \in \Omega } [ \ell_{\text{\IA}} ( \btheta; \mathbf x^\prime, \mathbf x) ], ~ 
    \ell_{\text{\IA}} ( \btheta; \mathbf x^\prime, \mathbf x) \Def    \underbrace{ {\mathrm{CE}}( \hat {f} (\mathbf x_{\mathrm{RED}}), \hat{f}(\mathbf x)) }_\text{{\IA} for benign prediction} + \underbrace{  {\mathrm{CE}}( \hat {f} (\mathbf x_{\mathrm{RED}}^{\prime}), \hat{f}(\mathbf x^{\prime})) }_\text{{\IA} for adversarial prediction}, 
\end{align}}%
where CE denotes the cross-entropy loss.
To enhance the class-discriminative ability, it is desirable to
integrate  the denoising loss \eqref{eq: MAE_only} with the PA regularization \eqref{eq: IA}, leading to $\ell_{\mathrm{denoise}} + \lambda  \ell_{\text{\IA}}$, where $\lambda > 0$ is a regularization parameter. 
To address this issue,  we will further propose a data augmentation method to improve the denoising ability  without losing the advantage of {\IA} regularization.

\paragraph{Proper data augmentation improves RED.}

\begin{wrapfigure}{r}{70mm}
\vspace*{-5mm}
\centering{
\begin{tabular}{c}
\hspace*{-3mm}
\includegraphics[width=0.48\textwidth]{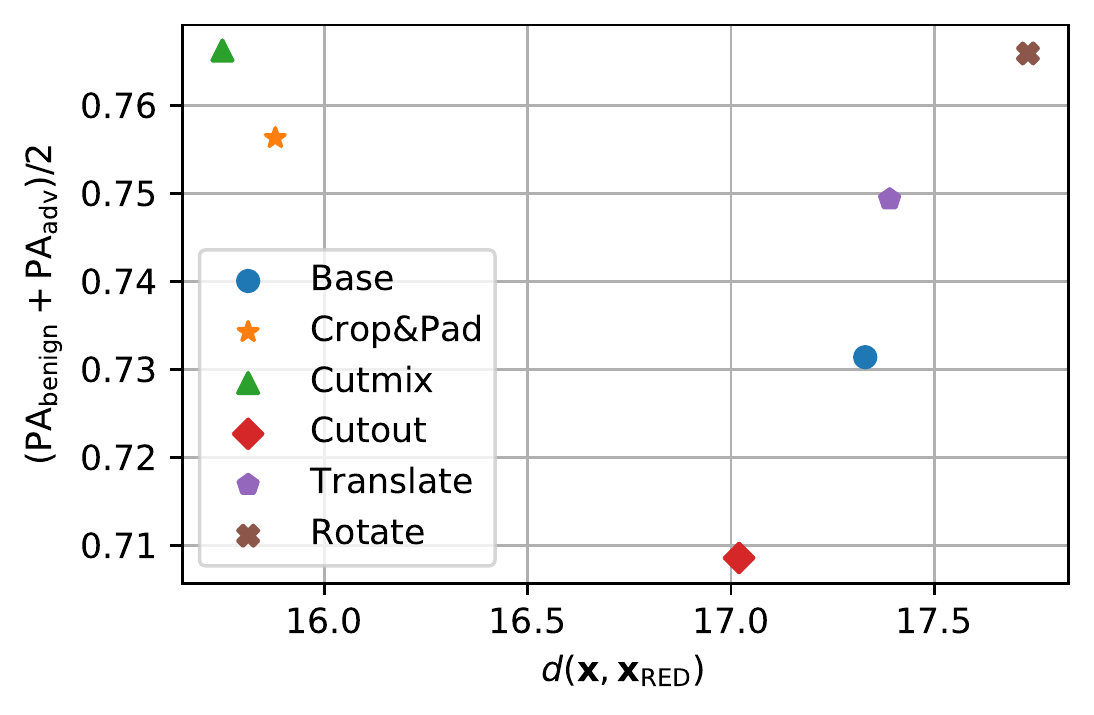}
\end{tabular}
}
\vspace*{-6mm}
\caption{\footnotesize{The influence of different data augmentations. `Base' refers to the base training without   augmentation.
}}
\vspace*{-5mm}
\label{fig: aug_study}
\end{wrapfigure}

The rationale behind 
 incorporating image transformations  into  {\DeRED} lies in two aspects.
First, data transformation  can
make RED foveated to the most informative attack artifacts since   an adversarial instance could be {sensitive} to  input transformations \citep{luo2015foveation,athalye18b,xie2019improving,li2020practical, fan2021does}.
Second,  the identification of  
  transformation-resilient benign/adversarial instances may enhance the capabilities of {\IA}  and {\AtrA}.
  
  However, it is highly non-trivial to determine the most appropriate   data augmentation operations. For example, a pixel-sensitive data transformation, e.g., Gaussian blurring and colorization, would hamper the reconstruction-ability of the original 
   adversary-benignity pair $(\mathbf x^\prime, \mathbf x)$. Therefore, we focus on spatial image transformations, including rotation, translation,  cropping \& padding, cutout, and CutMix \citep{yun2019cutmix}, which keep the original perturbation in a linear way. 
 In Fig.\ref{fig: aug_study}, we evaluate the RED performance, in terms of pixel-level reconstruction error  and  prediction-level alignment accuracy, for different kinds of spatial image transformations. As we can see, CutMix and  cropping \& padding can increase the both performance simultaneously, considered as the appropriate augmentation to boost the RED. Furthermore, we empirically find that combining the two transformations can further  improve the performance.

  Let $\mathcal T$ denote a transformation set, including cropping \& padding and CutMix operations. 
  With the aid of the denoising loss  \eqref{eq: MAE_only}, {\IA} regularization \eqref{eq: IA}, and data transformations $\mathcal T$,  we then cast
 the overall training objective of {\DeRED} as  

 {\small
 \vspace*{-3.5mm}
 \begin{align}
  \hspace*{-5mm}  \begin{array}{ll}
\displaystyle \minimize_{\btheta}         & 
\underbrace{ \mathbb E_{(\mathbf x^{\prime}, \mathbf x) \in \Omega , t \sim   \mathcal T }\|  \mathcal D_{\btheta} (t(\mathbf x^{\prime})) - t(\mathbf x) \|_1 }_\text{$\ell_{\mathrm{denoise}}$ \eqref{eq: MAE_only} with data augmentations}   + \underbrace{  \lambda
\mathbb E_{(\mathbf x^\prime, \mathbf x) \in \Omega, t \sim \check {\mathcal T} } [  
\ell_{\text{\IA}} ( \btheta; t(\mathbf x^\prime), t(\mathbf x))
] }_\text{$\ell_{\text{\IA}}$ \eqref{eq: IA} with data augmentation via $\check{\mathcal T}$},
    \end{array}
\hspace*{-5mm}    \label{eq: overall_DeRED}
\end{align}}%
where 
$\check{\mathcal T}$ denotes a properly-selected subset of 
$\mathcal T$, and $\lambda > 0$  is a regularization parameter. 
In the {\IA} regularizer  \eqref{eq: overall_DeRED}, we need to avoid the scenario of
    over-transformation   where  data augmentation alters the classifier's original decision. 
This suggests  
$ \check{\mathcal T} = \{ t \in \mathcal T\,| \,  \hat{F}(t(\mathbf{x})) = \hat{F}(\mathbf{x}),  \hat{F}(t(\mathbf{x}^{\prime})) = \hat{F}(\mathbf{x}^{\prime}) \ \}$, where $\hat{F}$ represents the prediction label of the pre-trained classifier $\hat f$, \textit{i.e.}, $\hat{F}(\cdot)=\mathrm{argmax}(\hat f(\cdot))$.

\section{Experiments}
\label{sec: exp}
We show the effectiveness of our proposed method in {$5$ aspects}: \textbf{a)} reconstruction error of adversarial perturbation inversion, \textit{i.e.,} $d(\mathbf x, \mathbf x_\mathrm{RED})$,
\textbf{b)} class-discriminative ability of the benign and adversarial example estimate, \textit{i.e.}, 
$\text{\IA}_{\mathrm{benign}}$ and $\text{\IA}_{\mathrm{adv}}$ by victim models,
\textbf{c)} adversary saliency region recovery, \textit{i.e.}, attribution alignment, and 
\textbf{d)} RED evaluation over unseen attack types and adaptive attacks.

\SubSection{Experiment setup}
\label{sec: exp_setup}

\paragraph{Attack datasets.}

To train and test RED models, we generate adversarial examples on the ImageNet dataset \citep{deng2009imagenet}.
We consider \textbf{$3$ attack methods} including PGD \citep{madry2017towards}, FGSM  \citep{Goodfellow2015explaining}, and CW attack \citep{carlini2017towards}, applied to \textbf{$5$ models} including pre-trained ResNet18 (Res18), Resnet50 (Res50) \citep{DBLP:journals/corr/HeZRS15}, VGG16, VGG19, and InceptionV3 (IncV3) \citep{DBLP:journals/corr/SzegedyVISW15}.  
{The detailed parameter settings can be found in Appendix \ref{sec:dataset_details}.} Furthermore, to evaluate the RED performance on unseen perturbation types during training, additional 2K adversarial examples generated by  \textbf{AutoAttack} \citep{croce2020reliable} and 1K adversarial examples generated by \textbf{Feature Attack} \citep{sabour2015adversarial}  are included as the unseen testing dataset. AutoAttack is applied on VGG19, Res50 and \textbf{two new victim models}, i.e., Alexet and Robust Resnet50 (R-Res50), via fast adversarial training \citep{wong2020fast} while Feature Attack is applied on VGG19 and Alexnet. The rational behind considering Feature Attack is that feature adversary has been recognized as an effective way to circumvent adversarial detection \citep{tramer2020adaptive}. Thus, it provides a supplement on detection-aware attacks. 

\paragraph{RED model configuration, training and evaluation.} During the training of the RED denoisers, VGG19 \citep{vgg19} is chosen as the pretrained classifier $\hat{f}$ for {\IA} regularization. Although different victim models were used for generating adversarial examples, we will show that the inference guided by VGG19 is able to accurately estimate the true image class and the intent of the adversary. In terms of the  architecture of $\mathcal D_{\btheta}$, DnCNN \citep{DNCNN_denoiser} is adopted. 
The RED problem is solved using an Adam optimizer \citep{KingmaB2015adam} with the initial learning rate of $10^{-4}$, which decays $10$ times for every $140$ training epochs. In \eqref{eq: overall_DeRED}, the regularization parameter $\lambda$ is set as $0.025$. The transformations for data augmentation include CutMix and cropping \& padding. The maximum number of training epochs  is set as $300$. 
{The computation cost and ablation study of {\DeRED} are in Appendix \ref{sec:cost} and \ref{sec:ablation}, respectively.}

\paragraph{Baselines.}
We compare {\DeRED} with two baseline approaches: \textbf{a)} the conventional denoising-only (DO) approach with the objective function \eqref{eq: MAE_only}; \textbf{b)} The state-of-the-art Denoised Smoothing (DS) \citep{salman2020denoised} approach that considers both the reconstruction error and the {\IA} for benign examples in the objective function.
Both methods are tuned to their best configurations. 

\subsection{Main  results}

\paragraph{Reconstruction error {$d(\mathbf x, \mathbf x_\mathrm{RED})$} and {\IA}.}
Table  \ref{table:RMSE_IA_comparison} presents the comparison of {\DeRED} with the baseline denoising approaches in terms of $d(\mathbf x, \mathbf x_\mathrm{RED})$, $d(f(\mathbf x), f(\mathbf x_\mathrm{RED}))$, $d(f(\mathbf x'), f(\mathbf x'_\mathrm{RED}))$, $\text{\IA}_{\mathrm{benign}}$, and $\text{\IA}_{\mathrm{adv}}$ on the testing dataset. As we can see, our approach ({\DeRED}) improves the class-discriminative ability from benign perspective by 42.91\% and adversarial perspective by 8.46\% with a slightly larger reconstruction error compared with the DO approach.
\begin{wraptable}{r}{58mm}
\begin{center}
\vspace*{-6mm}
\caption{\footnotesize{The performance comparison among DO, DS and {\DeRED} on the testing dataset.}} \label{table:RMSE_IA_comparison}
\begin{threeparttable}
\resizebox{0.4\textwidth}{!}{
\begin{tabular}{c|c|c|c}
\toprule\hline
                                       & DO      & DS      & \DeRED \\ \hline
$d(\mathbf x, \mathbf x_\mathrm{RED})$ & 9.32    & 19.19   & 13.04                 \\ \hline
$d(f(\mathbf x), f(\mathbf x_\mathrm{RED}))$ & 47.81   & 37.21   & 37.07                 \\ \hline
$d(f(\mathbf x'), f(\mathbf x'_\mathrm{RED}))$ & 115.09    &  150.02  &      78.21          \\ \hline
$\text{\IA}_{\mathrm{benign}}$         & 42.80\% & 86.64\% & 85.71\%               \\ \hline
$\text{\IA}_{\mathrm{adv}}$            & 71.97\% & 72.47\% & 80.43\%               \\ \hline\bottomrule
\end{tabular}}
\end{threeparttable}
\end{center}
\vspace*{-7mm}
\end{wraptable}
In contrast to DS, 
{\DeRED} achieves similar $\text{\IA}_{\mathrm{benign}}$ 
but improved pixel-level   denoising error and $\text{\IA}_{\mathrm{adv}}$. Furthermore, {\DeRED}  achieves the best logit-level reconstruction error for both $f(\mathbf x_\mathrm{RED})$ and $f(\mathbf x'_\mathrm{RED})$ among the three approaches. This implies that $\mathbf{x}_\mathrm{RED}$ rendered by {\DeRED} can achieve highly similar prediction to the true benign example $\mathbf x$, and the perturbation estimate ${\mathbf x^{\prime} -\mathbf x_{\mathrm{RED}}}$ yields a similar misclassification effect to the ground-truth perturbation. 
{Besides, {\DeRED} is robust against attacks with different hyperparameters settings, details can be found in Appendix \ref{sec:diff_attack_hyper}.}

 \begin{figure*}[htb]
  \centering
  \begin{adjustbox}{max width=1\textwidth }
  \begin{tabular}{@{\hskip 0.00in}c @{\hskip 0.00in}c | @{\hskip 0.00in} @{\hskip 0.02in} c @{\hskip 0.02in} | @{\hskip 0.02in} c @{\hskip 0.02in} |@{\hskip 0.02in} c @{\hskip 0.02in} |@{\hskip 0.02in} c @{\hskip 0.02in} 
  }
& 
\colorbox{lightgray}{ \textbf{Input image}}
&
\colorbox{lightgray}{ \textbf{DO}} 
&  
\colorbox{lightgray}{ \textbf{DS}} 
&
\colorbox{lightgray}{  \textbf{{\DeRED} (ours)}}
&
\colorbox{lightgray}{  \textbf{Groundtruth}}

\\
 \begin{tabular}{@{}c@{}}  

\rotatebox{90}{\parbox{10em}{\centering  \textbf{\large{Benign example $\mathbf x$/$\mathbf x_{\mathrm{RED}}$}}}}
 \\

\rotatebox{90}{\parbox{10em}{\centering  \textbf{\large{Adv. example $\mathbf x^\prime$/$\mathbf x_{\mathrm{RED}}^\prime$} 
}}}

\end{tabular} 
& 
\begin{tabular}{@{\hskip 0.02in}c@{\hskip 0.02in}}
\\
 \begin{tabular}{@{\hskip 0.02in}c@{\hskip 0.02in} }
 \parbox[c]{10em}{\includegraphics[width=10em]{Figs/piggybank/clean.jpg}

 }  
\end{tabular}
 \\
 \begin{tabular}{@{\hskip 0.02in}c@{\hskip 0.02in}}
 \parbox[c]{10em}{\includegraphics[width=10em]{Figs/piggybank/adv.jpg}}  
\end{tabular}
\end{tabular}
&
\begin{tabular}{@{\hskip 0.02in}c@{\hskip 0.02in}}
 \begin{tabular}{@{\hskip 0.02in}c@{\hskip 0.02in}c@{\hskip 0.02in} }
      \begin{tabular}{@{\hskip 0.00in}c@{\hskip 0.00in}}
     \parbox{10em}{\centering   $I(\cdot, y)$}  
    \end{tabular} 
     &  
      \begin{tabular}{@{\hskip 0.00in}c@{\hskip 0.00in}}
       \parbox{10em}{\centering   $I(\cdot, y^\prime )$} 
    \end{tabular} 
\end{tabular}\\
 \begin{tabular}{@{\hskip 0.02in}c@{\hskip 0.02in}c@{\hskip 0.02in} }
 \parbox[c]{10em}{\includegraphics[width=10em]{Figs/piggybank/DOcc.jpg}

 } &    \parbox[c]{10em}{\includegraphics[width=10em]{Figs/piggybank/DOct.jpg}} 
\end{tabular}
 \\
 \begin{tabular}{@{\hskip 0.02in}c@{\hskip 0.02in}c@{\hskip 0.02in} }
 \parbox[c]{10em}{\includegraphics[width=10em]{Figs/piggybank/DOac.jpg}} &    \parbox[c]{10em}{\includegraphics[width=10em]{Figs/piggybank/DOat.jpg}} \\
\end{tabular}
\end{tabular}
&
\begin{tabular}{@{\hskip 0.02in}c@{\hskip 0.02in}}
 \begin{tabular}{@{\hskip 0.02in}c@{\hskip 0.02in}c@{\hskip 0.02in} }
      \begin{tabular}{@{\hskip 0.00in}c@{\hskip 0.00in}}
     \parbox{10em}{\centering   $I(\cdot, y)$}  
    \end{tabular} 
     &  
      \begin{tabular}{@{\hskip 0.00in}c@{\hskip 0.00in}}
       \parbox{10em}{\centering   $I(\cdot, y^\prime )$} 
    \end{tabular} 
\end{tabular}\\
 \begin{tabular}{@{\hskip 0.02in}c@{\hskip 0.02in}c@{\hskip 0.02in} }
 \parbox[c]{10em}{\includegraphics[width=10em]{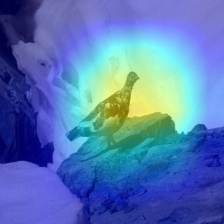}} &    \parbox[c]{10em}{\includegraphics[width=10em]{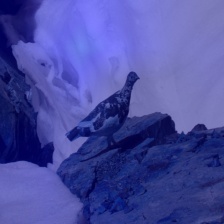}} 
\end{tabular}
 \\
 \begin{tabular}{@{\hskip 0.02in}c@{\hskip 0.02in}c@{\hskip 0.02in} }
 \parbox[c]{10em}{\includegraphics[width=10em]{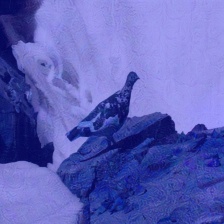}} &    \parbox[c]{10em}{\includegraphics[width=10em]{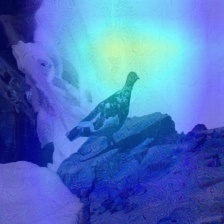}} 
 \\
\end{tabular}
\end{tabular}
&
\begin{tabular}{@{\hskip 0.02in}c@{\hskip 0.02in}}
 \begin{tabular}{@{\hskip 0.02in}c@{\hskip 0.02in}c@{\hskip 0.02in} }
      \begin{tabular}{@{\hskip 0.00in}c@{\hskip 0.00in}}
     \parbox{10em}{\centering   $I(\cdot, y)$}  
    \end{tabular} 
     &  
      \begin{tabular}{@{\hskip 0.00in}c@{\hskip 0.00in}}
       \parbox{10em}{\centering   $I(\cdot, y^\prime )$} 
    \end{tabular} 
\end{tabular}\\
 \begin{tabular}{@{\hskip 0.02in}c@{\hskip 0.02in}c@{\hskip 0.02in} }
 \parbox[c]{10em}{\includegraphics[width=10em]{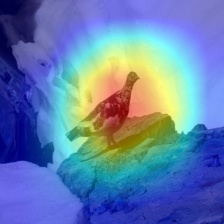}} &    \parbox[c]{10em}{\includegraphics[width=10em]{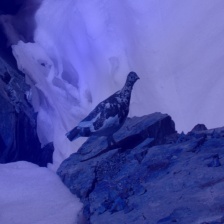}} 
\end{tabular}
 \\
 \begin{tabular}{@{\hskip 0.02in}c@{\hskip 0.02in}c@{\hskip 0.02in} }
 \parbox[c]{10em}{\includegraphics[width=10em]{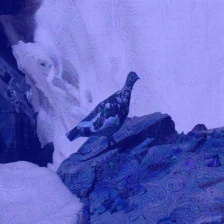}} &    \parbox[c]{10em}{\includegraphics[width=10em]{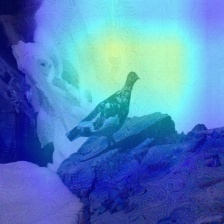}} 
 \\
\end{tabular}
\end{tabular}
&
\begin{tabular}{@{\hskip 0.02in}c@{\hskip 0.02in}}
 \begin{tabular}{@{\hskip 0.02in}c@{\hskip 0.02in}c@{\hskip 0.02in} }
      \begin{tabular}{@{\hskip 0.00in}c@{\hskip 0.00in}}
     \parbox{10em}{\centering   $I(\cdot, y)$}  
    \end{tabular} 
     &  
      \begin{tabular}{@{\hskip 0.00in}c@{\hskip 0.00in}}
       \parbox{10em}{\centering   $I(\cdot, y^\prime )$} 
    \end{tabular} 
\end{tabular}\\
 \begin{tabular}{@{\hskip 0.02in}c@{\hskip 0.02in}c@{\hskip 0.02in} }
 \parbox[c]{10em}{\includegraphics[width=10em]{Figs/piggybank/cc_gt.jpg}} &    \parbox[c]{10em}{\includegraphics[width=10em]{Figs/piggybank/ct_gt.jpg}} 
\end{tabular}
 \\
 \begin{tabular}{@{\hskip 0.02in}c@{\hskip 0.02in}c@{\hskip 0.02in} }
 \parbox[c]{10em}{\includegraphics[width=10em]{Figs/piggybank/ac_gt.jpg}} &    \parbox[c]{10em}{\includegraphics[width=10em]{Figs/piggybank/at_gt.jpg}} 
 \\
\end{tabular}
\end{tabular}

\end{tabular}
  \end{adjustbox}
\caption{\footnotesize{
Interpretation  ($I$) of benign  ($\mathbf x$/$\mathbf x_{\mathrm{RED}}$) and adversarial ($\mathbf x^\prime$/$\mathbf x_{\mathrm{RED}}^\prime$)    image w.r.t. the true label {$y$=`ptarmigan'} and the adversary targeted label {$y^\prime$=`shower curtain'}.
We compare three methods of RED training,  DO, DS, and {\DeRED} as our method, to the ground-truth interpretation.  Given an RED method,  the first column is  $I(\mathbf x_{\mathrm{RED}}, y)$ versus $I(\mathbf x_{\mathrm{RED}}^\prime , y)$,   the second column is $I(\mathbf x_{\mathrm{RED}}, y^\prime)$ versus $I(\mathbf x_{\mathrm{RED}}^\prime , y^\prime)$, and  all maps under each RED method are normalized w.r.t. their largest value. For the ground-truth, the first column is  $I(\mathbf x, y)$ versus $I(\mathbf x^\prime , y)$,   the second column is $I(\mathbf x, y^\prime)$ versus $I(\mathbf x^\prime , y^\prime)$.
}}
\label{fig: attribution}
\end{figure*}

\begin{figure}[htb]
\centering
\subfigure
[Denoising Only] 
{\includegraphics[width=12em]{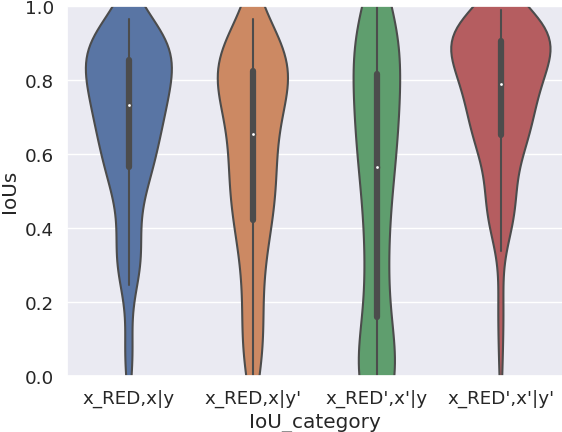}} 
\subfigure[Denoised Smoothing]{\includegraphics[width=12em]{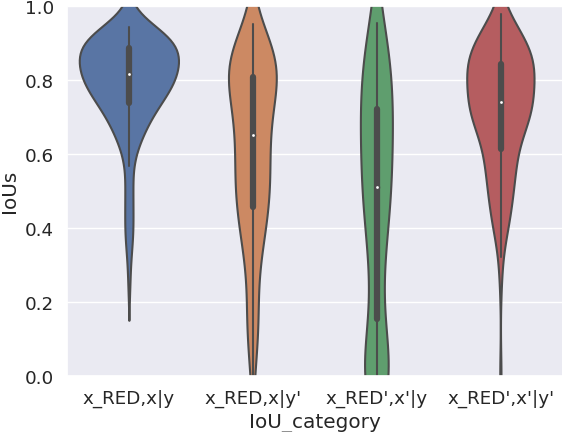}}
\subfigure[{\DeRED} (ours)]{\includegraphics[width=12em]{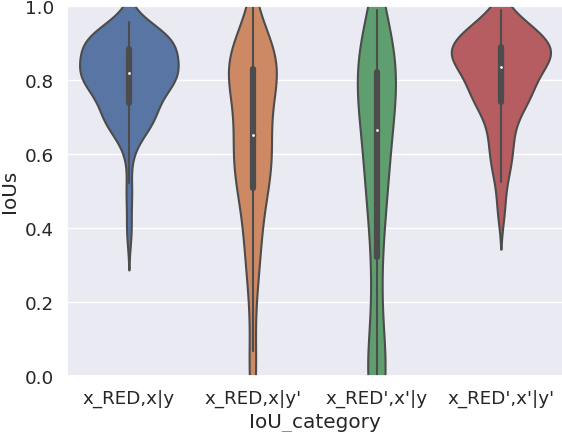}}
\caption{\footnotesize{IoU distributions of the attribution alignment by three RED methods. Higher IoU is better. For each subfigure, the four IoU scores standing for $\mathrm{IoU}( \mathbf x_{\mathrm{RED}}, \mathbf x, y )$,
$\mathrm{IoU}( \mathbf x_{\mathrm{RED}}, \mathbf x, y^\prime )$,
$\mathrm{IoU}( \mathbf x_{\mathrm{RED}}^\prime, \mathbf x^\prime, y )$,
and
$\mathrm{IoU}( \mathbf x_{\mathrm{RED}}^\prime, \mathbf x^\prime, y^\prime )$.} }
\label{fig: IoU} 
\vspace*{-3mm}
\end{figure}

\paragraph{Attribution alignment.}
In addition to pixel-level alignment and prediction-level alignment to evaluate the RED performance, attribution alignment is examined in what follows. 
Fig.\,\ref{fig: attribution} presents attribution maps generated by GradCAM 
{in terms of $ I(\mathbf x, y)$, $ I(\mathbf x^\prime, y)$, $I(\mathbf x, y^\prime)$, and $ I(\mathbf x^\prime, y^\prime)$, where  $\mathbf x^\prime$ denotes the perturbed version of $\mathbf x$, and $y^\prime$ is the adversarially targeted label}. From left to right is the attribution map over DO, DS, {\DeRED} (our method), and the ground-truth. 
Compared with DO and DS, {\DeRED} yields 
a closer attribution alignment with the ground-truth especially when making a comparison between $I(\mathbf x_{\mathrm{RED}}, y)$ and $I(\mathbf x, y)$.
At the dataset level, 
Fig.\,\ref{fig: IoU} shows the distribution of attribution IoU scores. It is observed that the IoU distribution of {\DeRED}, compared with DO and DS, has a denser concentration over the high-value area, corresponding to closer alignment with the attribution map by the adversary. {This feature indicates an interesting application of the proposed RED approach, which is to achieve the recovery of adversary's saliency region, in terms of the class-discriminative image regions that the adversary focused on.}

\vspace{-2mm}
\paragraph{RED vs. unforeseen attack types.}

\begin{figure}[htb]
\centering
\subfigure
[Accuracy of $\mathbf{x}^{\prime p}_{\mathrm{RED}}$] 
{\includegraphics[width=0.33\linewidth]{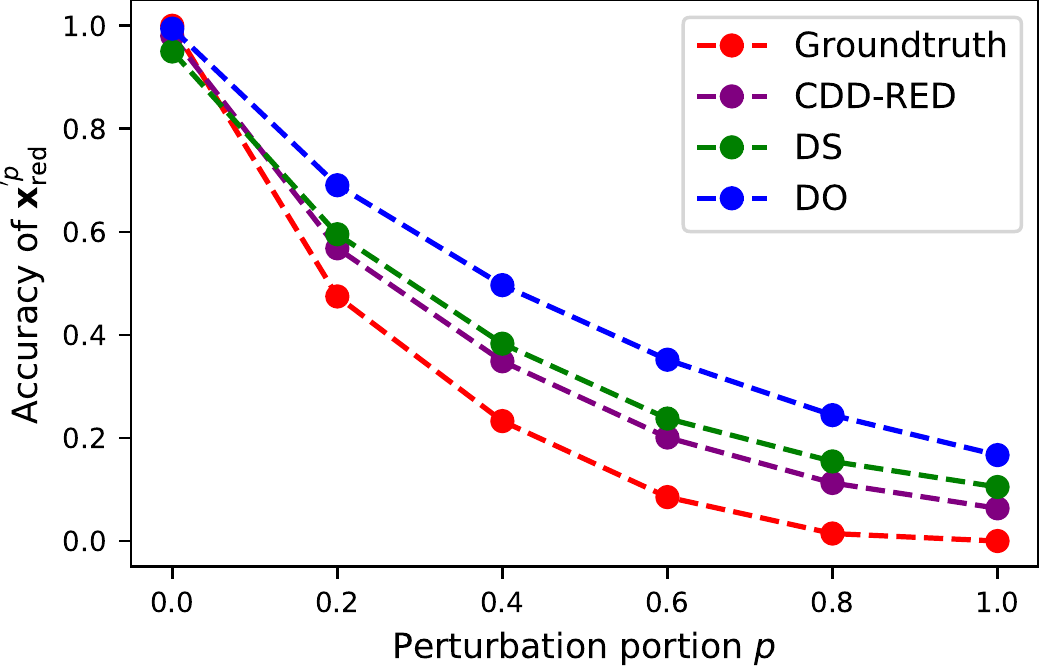}} 
\subfigure[Success rate of $\mathbf{x}^{\prime p}_{\mathrm{RED}}$]{\includegraphics[width=0.33\linewidth]{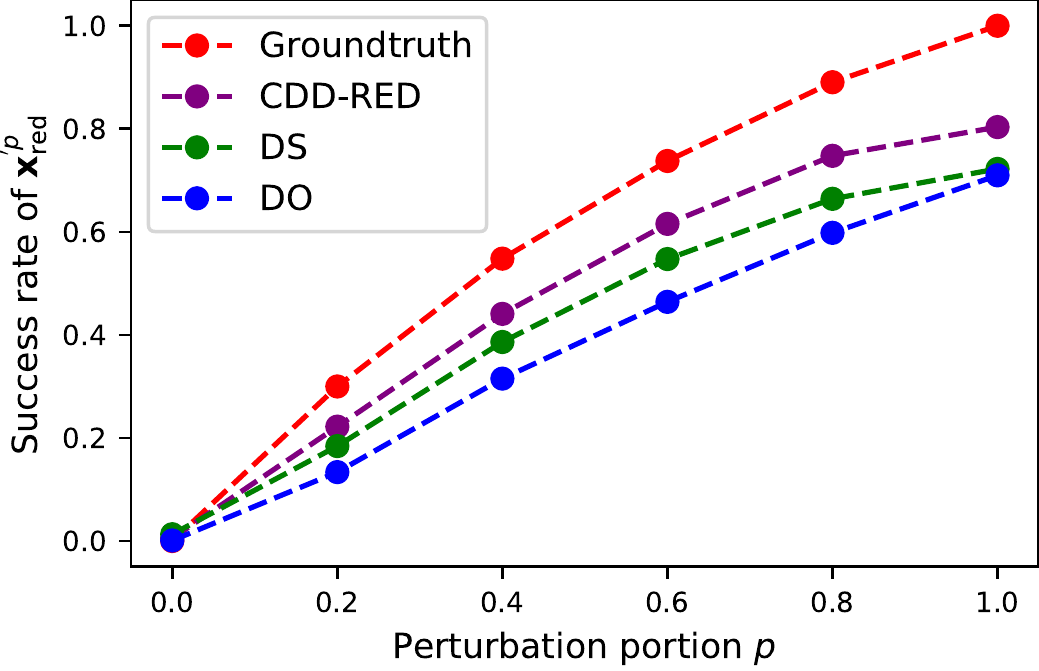}}
\subfigure[$d(f(\mathbf{x}'^{p}_{\mathrm{RED}}),f(\mathbf{x}))$]{\includegraphics[width=0.332\linewidth]{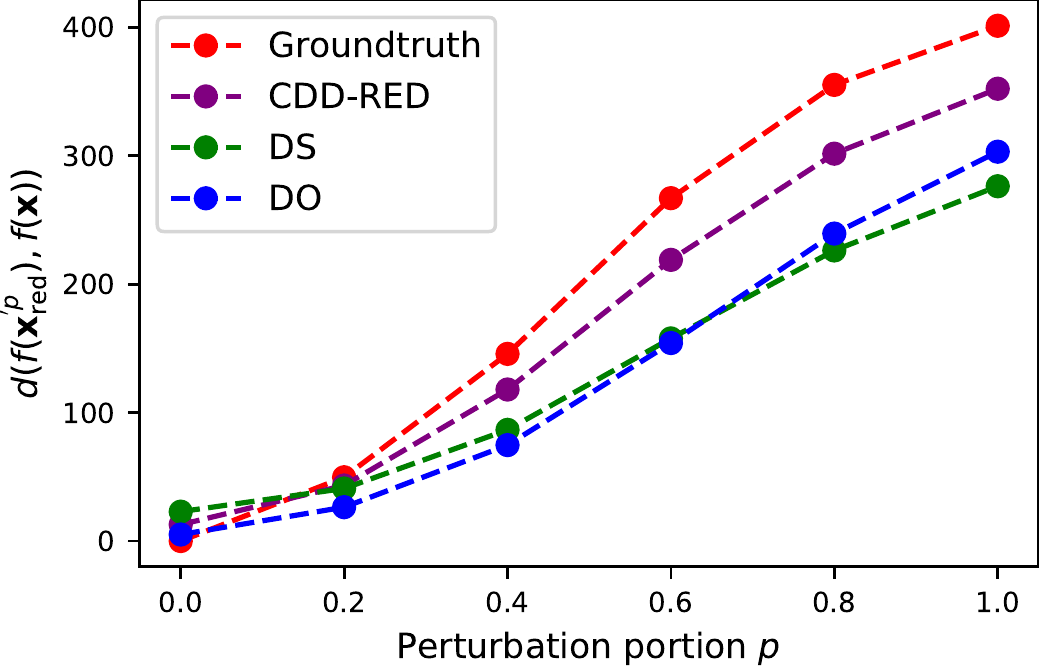}}
\subfigure[$d(\mathbf{x}'^{p}_{\mathrm{RED}},\mathbf{x})$]{\includegraphics[width=0.33\linewidth]{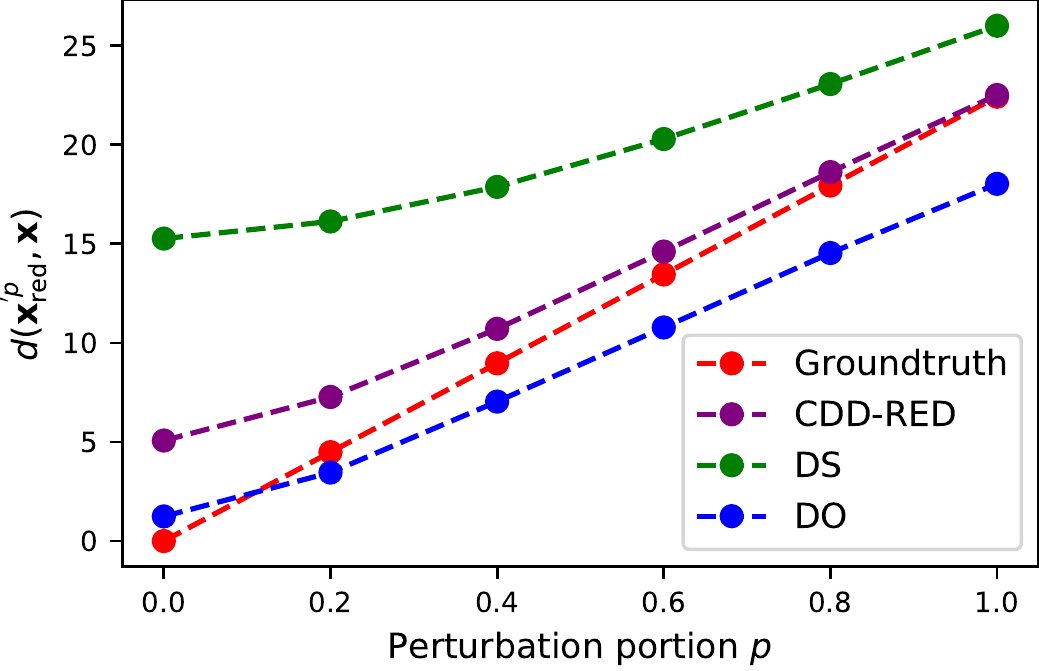}}
\vspace{-5mm}
\caption{\footnotesize{Reverse engineer partially-perturbed data under different interpolation portion $p$.} }
\vspace{-7mm}
  \label{fig:interpolation}  
\end{figure}

The experiments on the recovery of unforeseen attack types  are composed of two parts: \textbf{a)} partially-perturbed data via linear interpolation, and \textbf{b)} the unseen attack type,  AutoAttack, Feature Attack, and Adaptive Attack. 
{More   attack results including adversarial attacks on CIFAR-10 dataset, Wasserstein minimized attacks, and attacks on smoothed classifiers can be found in Appendix \ref{sec:appendix_unseen}.}

First, we construct  partially-perturbed   data by adding a portion $p \in \{0\%, 20\%, \cdots, 100\%\}$ of the perturbation $\mathbf{x}^{\prime}-\mathbf{x}$ to the true benign example $\mathbf{x}$, namely, $\mathbf{x}^{\prime p} = \mathbf{x} + p(\mathbf{x}^{\prime}-\mathbf{x})$. The interpolated $\mathbf{x}^{\prime p}$ is then used as the input to an RED model. We aim to investigate whether or not the proposed RED method can recover partial perturbations (even not successful attacks). 

Fig.\,\ref{fig:interpolation} (a) and (b) show the the prediction alignment with $y$ and $y^\prime$, of the adversarial example estimate $\mathbf{x}^{\prime p}_{\mathrm{RED}}=\mathbf{x}^{\prime p}-\mathcal D_{\btheta}(\mathbf{x}^{\prime p})+\mathbf{x}$ by different RED models. Fig.\,\ref{fig:interpolation} (c) shows the logit distance between the prediction of the partially-perturbed adversarial example estimate and the prediction of the benign example while Fig.\,\ref{fig:interpolation} (d) demonstrates the pixel distance between $\mathbf{x}^{\prime p}_{\mathrm{RED}}$ and the benign example. 

First, a smaller gap between the ground-truth curve (in red) and the adversarial example estimate $\mathbf{x}^{\prime p}_{\mathrm{red}}$ curve indicates a better performance. 
Fig.\,\ref{fig:interpolation} (a) and (b) show that {\DeRED} estimates the closest adversary's performance to the ground truth in terms of the prediction accuracy and attack success rate. This is also verified by the distance of prediction logits in Fig.\,\ref{fig:interpolation} (c). Fig.\,\ref{fig:interpolation} (d) 
shows that DS largely over-estimates the additive perturbation, while {\DeRED} maintains the perturbation estimation performance closest to the ground truth. {Though DO is closer to the ground-truth than {\DeRED} at p < 40\%, DO is not able to recover a more precise adversarial perturbation in terms of other performance metrics. For example, in Fig.\,\ref{fig:interpolation} (b) at p = 0.2, $\mathbf{x}^{\prime p}_{\mathrm{RED}}$ by DO achieves a lower successful attack rate compared to CDD-RED and the ground-truth.}

\begin{wraptable}{r}{80mm}
\begin{center}
\vspace*{-8mm}
\caption{\footnotesize{The $d(\mathbf x, \mathbf x_\mathrm{RED})$, $\text{\IA}_{\mathrm{benign}}$, and $\text{\IA}_{\mathrm{adv}}$ performance of the denoisers on the unforeseen perturbation type AutoAttack, Feature Attack, and Adaptive Attack.
}}
\label{table:unforeseen_results}
\begin{threeparttable}
\resizebox{0.57\textwidth}{!}{
\begin{tabular}{c|c|c|c|c}
\toprule\hline
\multicolumn{2}{c|}{}                                                    & DO      & DS      & \DeRED \\ \hline
\multirow{3}{*}{$d(\mathbf x, \mathbf x_\mathrm{RED})$} & AutoAttack    & 6.41    & 16.64   & 8.81                  \\ \cline{2-5} 
                                                        & Feature Attack & 5.51    & 16.14   & 7.99                  \\ \cline{2-5} 
                                                        & Adaptive Attack& 9.76    & 16.21   & 12.24                 \\ \hline         
\multirow{3}{*}{$\text{\IA}_{\mathrm{benign}}$}         & AutoAttack    & 84.69\% & 92.64\% & 94.58\%               \\ \cline{2-5} 
                                                        & Feature Attack & 82.90\% & 90.75\% & 93.25\%               \\ \cline{2-5} 
                                                        & Adaptive Attack& 33.20\% & 27.27\% & 36.29\%               \\ \hline
\multirow{3}{*}{$\text{\IA}_{\mathrm{adv}}$}            & AutoAttack    & 85.53\% & 83.30\% & 88.39\%               \\ \cline{2-5} 
                                                        & Feature Attack & 26.97\% & 35.84\% & 63.48\%               \\ \cline{2-5} 
                                                        & Adaptive Attack& 51.21\% & 55.41\% & 57.11\%               \\ \hline\bottomrule
\end{tabular}}
\end{threeparttable}
\end{center}
\vspace{-5mm}
\end{wraptable}
Moreover, as for benign examples with $p=0\%$ perturbations, though the RED denoiser does not see benign example pair ($\mathbf x$, $\mathbf x$) during training, it keeps the performance of the benign example recovery. {\DeRED} can handle the case with a mixture of adversarial and benign examples. That is to say, even if a benign example, detected as adversarial, is wrongly fed into the RED framework, our method can recover the original perturbation close to the ground truth. {See Appendix \ref{sec:mixture} for details.} 

Table\,\ref{table:unforeseen_results} shows the RED performance  on the unseen attack type,  AutoAttack, Feature Attack, and Adaptive Attack.  For AutoAttack and Feature Attack, {\DeRED} outperforms both DO and DS in terms of PA from both benign and adversarial perspectives. Specifically, {\DeRED} increases the $\text{\IA}_{\mathrm{adv}}$ for Feature Attack by 36.51\% and 27.64\% compared to DO and DS, respectively. 

As for the adaptive attack {\citep{tramer2020adaptive}}, we assume that the attacker has access to the knowledge of the RED model, \textit{i.e.}, $D_{\btheta}$. It can then perform the PGD attack method to generate successful prediction-evasion attacks even after taking the RED operation.

We use PGD methods to generate such  attacks  within the $\ell_\infty$-ball of 
perturbation radius 
$\epsilon=20/255$. Table\,\ref{table:unforeseen_results} shows that Adaptive Attack is much stronger than Feature Attack and AutoAttack, leading to larger reconstruction error and lower PA. However,  {\DeRED} still outperforms DO and DS in    $\text{\IA}_{\mathrm{benign}}$ and $\text{\IA}_{\mathrm{adv}}$. Compared to DS, it achieves a better trade-off with denoising error $d(\mathbf x, \mathbf x_\mathrm{RED})$.

In general, {\DeRED} can achieve high PA even for unseen attacks, indicating the generalization-ability of our method to estimate not only new adversarial examples (generated from the same attack method), but also new attack types.
\vspace{-3mm}
\paragraph{RED to infer correlation between adversaries.}

In what follows, we investigate whether the RED model guided by the single classifier (VGG19) enables to identify different adversary classes, given by combinations of attack types (FGSM, PGD, CW) and   victim model types (Res18, Res50, VGG16, VGG19, IncV3). 

\begin{wrapfigure}{r}{100mm}
\vspace{-10mm}
\centering
\subfigure
[Groundtruth]   
{\includegraphics[width=14em]{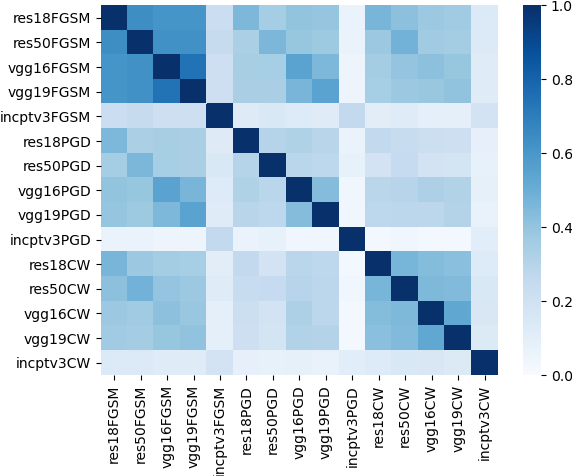}} 
\subfigure[{\DeRED} (ours)]{\includegraphics[width=14em]{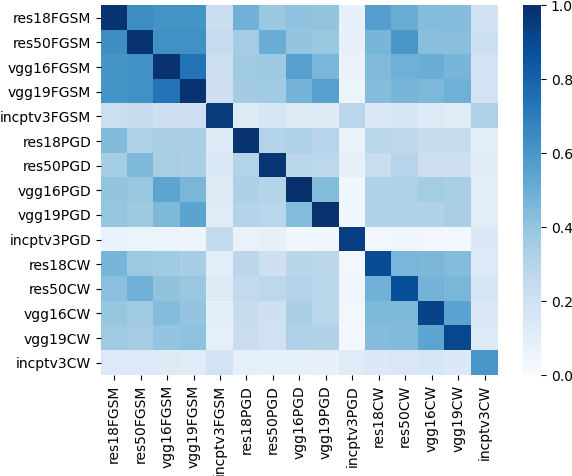}}
\vspace*{-5mm}
\caption{\footnotesize{Correlation matrices between different adversaries. For each correlation matrix, rows and columns represent adversarial example estimate $\mathbf x_{\mathrm{RED}}^\prime$ and true adversarial example $\mathbf x^\prime$ (For the ground-truth correlation matrix,   $\mathbf x_{\mathrm{RED}}^\prime = \mathbf x^\prime $). Each entry represents the average Spearman rank correlation between the logits of two adversary settings $\in$ \{(victim model, attack type)\}.
}}
\vspace*{-3mm}
\label{fig:correlationmatrix} 
\end{wrapfigure}
Fig.\,\ref{fig:correlationmatrix} presents 
the correlation between every two adversary classes in the logit space.  
Fig.\,\ref{fig:correlationmatrix} (a) shows the ground-truth correlation map. 
Fig.\,\ref{fig:correlationmatrix} (b) shows correlations  between logits of  $\mathbf x_{\mathrm{RED}}^\prime$ estimated by our RED  method ({\DeRED})
and logits of the true $\mathbf x^\prime$. 
Along the diagonal of each correlation matrix, the darker implies the better RED estimation under the same adversary class. 
By peering into off-diagonal entries, 
we find that   FGSM attacks
are more resilient to the choice of a victim model (see the cluster of high correlation values  at the top left corner of Fig.\,\ref{fig:correlationmatrix}).
Meanwhile, the proposed {\DeRED} precisely  recovers the correlation behavior of the true adversaries. Such a correlation matrix can help explain the similarities between different attacks' properties. Given an inventory of existing attack types, if a new attack appears, then one can resort to RED to estimate the correlations between the new attack type and the existing attack types. {Based on the correlation screening, it can infer the properties of the new attack type based on its most similar counterpart in the existing attack library; see Appendix \ref{sec:identity}. Inspired by the correlation, RED-synthesized perturbations can also be used as a transfer attack as well; see Appendix \ref{sec:tranferability}.} 
\vspace{-3mm}
\paragraph{Other applications of RED.}
{In Appendix \ref{section:potential_app}, we also empirically show that  the proposed RED approach can be applied  to adversarial detection, attack identity inference, and adversarial defense.}

\vspace*{-4mm}
\section{Conclusion}\label{ref:conclusion}
\vspace*{-3mm}
In this paper, we study the problem of Reverse Engineering of Deceptions (RED), to recover the attack signatures (\textit{e.g.} adversarial perturbations and adversary saliency regions) from an adversarial instance.  
To the best of our knowledge, RED has not been well studied. 
Our work makes a solid step towards formalizing the RED problem and developing a systematic pipeline, covering not only a solution but also a complete set of evaluation metrics. 
We have identified a series of RED principles, ranging from the pixel level to the attribution level, desired to reverse-engineer adversarial attacks. 
We have developed  an effective denoiser-assisted RED approach by integrating class-discrimination   and   data augmentation into an image denoising network. 
With extensive experiments, our approach outperforms the existing baseline methods and generalizes well to unseen attack types.

\section*{Acknowledgment}

The work is supported by the DARPA RED program. 
We also thank the MIT-IBM Watson AI Lab, IBM Research for supporting us with GPU computing resources.


{{
\bibliographystyle{iclr2022_conference}
\bibliography{refs,reference, ref_revision}
}}

\clearpage
\appendix
\setcounter{table}{0}  
\setcounter{figure}{0}

\renewcommand{\thetable}{A\arabic{table}}
\renewcommand{\thefigure}{A\arabic{figure}}

\section{RED by PGD attack: Performing PGD back to the true class} \label{sec:naive_red}

A naive approach to reverse engineer the adversarial perturbation is using the target PGD attack to revert the label back to the groundtruth. However, this requires additional assumptions. 
First, since PGD is a test-time deterministic optimization approach for perturbation generation, its targeted implementation requires the true class of the adversarial example, which could be unknown at testing time. What is more, one has to pre-define the perturbation budget $\epsilon$ for PGD. This value is also unknown. 
Second, performing PGD back to the true class might not exactly recover the ground-truth adversarial perturbations. By contrast, its RED counterpart could be over-perturbed. To make it more convincing, we applied the target $l_\infty$ PGD attack method to adversarial examples generated by PGD (assuming true class, victim model, and attack budget are known). We tried various PGD settings ($\text{PGD10}_{\epsilon=10/255}$ refers to PGD attack using 10 steps and $\epsilon=10/255$). Eventually, we compare these results to our CDD-RED method in Table \ref{table:PGD_back}.

Given that the average reconstruction error between $\mathbf x$ and $\mathbf x'$ is 20.60, we can see from Table \ref{table:PGD_back} that PGD attacks further enlarge the distortion from the clean data. Although PGD attacks can achieve high accuracy after reverting the adversarial data back to their true labels, the resulting perturbation estimate is far from the ground-truth in terms of their prediction alignment. We can tell from the low $\text{PA}_{\text{adv}}$  by PGD methods that $\mathbf x_{\mathrm {RED}}'$ does not align with the input $\mathbf x'$ at all.

\begin{table}[H]
\begin{center}
\caption{The performance comparison among DO, DS, and {\DeRED} on the CIDAR-10 dataset.} \label{table:PGD_back}
{
\begin{tabular}{ccccc}
\toprule\hline
                                       & PGD10{$\epsilon_{20/255}$} & PGD10{$\epsilon_{10/255}$}            & PGD20{$\epsilon_{20/255}$}            & CDD-RED                                                \\ \hline
$d(\mathbf x, \mathbf x_\mathrm{RED})$ & 27.63                                     & 22.67                                                 & 27.53                                                 & \textbf{11.73}                                                  \\ \hline
$\text{\IA}_{\mathrm{benign}}$         & 96.20\%                                & 82.60\%                                               & \textbf{99.80\%}                                               & 83.20\%                                                \\ \hline
$\text{\IA}_{\mathrm{adv}}$            & 6.20\%                                    & 7.20\% &  4.80\% & \textbf{97.40\%}\\ \hline\bottomrule
\end{tabular}}
\end{center}
\end{table}

\section{Dataset Details.} \label{sec:dataset_details}

The training and testing dataset is composed of three attack methods including  PGD \cite{madry2017towards}, FGSM  \cite{Goodfellow2015explaining}, and CW attack \cite{carlini2017towards}, applied to \textbf{5 models} including pre-trained ResNet18 (Res18), Resnet50 (Res50) \cite{DBLP:journals/corr/HeZRS15}, VGG16, VGG19, and InceptionV3 (IncV3) \cite{DBLP:journals/corr/SzegedyVISW15}. By default, PGD attack and FGSM attack are bounded by $\ell_\infty$-norm constraint and CW attack is bounded by $\ell_2$-norm. The range of the perturbation strength $\epsilon$ for PGD attack and FGSM attack are $[1/255,40/255)$ and $[1/255,20/255)$, respectively. As for CW attack, the adversary's confidence parameter $k$ is uniformly sampled from from $[1,20]$.
One attack method is applied to one victim model to obtain \textbf{3K successfully attacked images}. As a consequence, 45K ($3\times5\times3$K) adversarial images are generated in total: 37.5K for training and 7.5K for validating. 
The testing set contains 28K adversarial images generated with the same attack method \& victim model.

\section{performance on more   attack types} \label{sec:appendix_unseen}

{We show more evaluations of the RED approaches on unforeseen attacks during training. The denoisers are all trained on the training dataset containing adversarial examples generated on the ImageNet dataset, as in Appendix \ref{sec:dataset_details}. The test data includes adversarial examples generated on the CIFAR-10 dataset in \ref{sec:cifar10}, Wasserstein minimized attackers in \ref{sec:W_attack}, and attacks on smoothed classifiers in \ref{sec:attack_against_smoothed_classifiers}.}

\subsection{{Performance on CIFAR-10 dataset}} \label{sec:cifar10}

{
We further evaluate the performance of the RED approaches on the adversarial examples generated on the CIFAR-10 dataset. As the denoiser is input agnostic, we directly test the denoiser trained on adversarial examples generated on the ImageNet dataset. Here we consider the $10$-step PGD-$l_{\inf}$ attack generation method with the perturbation radius $\epsilon = 8/255$. And these examples are not seen during our training. As shown in the Table \ref{table:cifar10_result}, the proposed {\DeRED} method provides the best $\text{PA}_{\text{clean}}$ and $\text{PA}_{\text{adv}}$ with a slightly larger $d(\mathbf x, \mathbf x_\mathrm{RED})$ than DO. This is not surprising as DO focuses only on the pixel-level denoising error metric. However, as illustrated in Sec.\,\ref{sec:evaluation_metric}, the other metric like PA also plays a key role in evaluating the RED performance.
}

\begin{table}[H]
\begin{center}
\caption{{The performance comparison among DO, DS, and {\DeRED} on the CIDAR-10 dataset.}} \label{table:cifar10_result}
{
\begin{tabular}{c|c|c|c}
\toprule\hline
                                       & DO      & DS      & \DeRED \\ \hline
$d(\mathbf x, \mathbf x_\mathrm{RED})$ & \textbf{0.94}    & 4.50   & 1.52                 \\ \hline
$\text{\IA}_{\mathrm{benign}}$         & 9.90\% & 71.75\% & \textbf{71.80\%}               \\ \hline
$\text{\IA}_{\mathrm{adv}}$            & 92.55\% & 89.70\% & \textbf{99.55\%}               \\ \hline\bottomrule
\end{tabular}}
\end{center}
\end{table}

\subsection{{Performance on Wasserstein minimized attackers}} \label{sec:W_attack}

{
We further show the performance on Wasserstein minimized attackers, which is an unseen attack type during training. The adversarial examples are generated on the ImageNet sub-dataset using Wasserstein ball. 
We follow the same setting from  \cite{wong2019wasserstein}, where the attack radius $\epsilon$ is 0.1 and the maximum iteration is 400 under $l_\infty$ norm inside Wasserstein ball. 
The results are shown in Table \ref{table:w_attack}.
As we can see, Wasserstein attack is a more challenging attack type for RED than the $l_p$ attack types considered in the paper, justified by the lower prediction alignement $\text{PA}_\text{benign}$ across all methods. 
This implies a possible limitation of supervised training over ($l_2$ or $l_\infty$) attacks. One simple solution is to expand the training dataset using more diversified attacks (including Wasserstein attacks). However, we believe that the further improvement of the generalization ability of RED deserves a more careful study in the future, e.g., an extension from the supervised learning paradigm to the (self-supervised) pre-training and finetuning paradigm.
}

\begin{table}[H]
\begin{center}
\caption{{The performance comparison among DO, DS, and {\DeRED} on  Wasserstein minimized attackers.}} \label{table:w_attack}
{
\begin{tabular}{c|c|c|c}
\toprule\hline
                                       & DO      & DS      & \DeRED \\ \hline
$d(\mathbf x, \mathbf x_\mathrm{RED})$ & \textbf{9.79}    & 17.38   & 11.66                 \\ \hline
$\text{\IA}_{\mathrm{benign}}$         & 92.50\%		 & 96.20\% & \textbf{97.50\%}               \\ \hline
$\text{\IA}_{\mathrm{adv}}$            & 35.00\% & 	37.10\%	& \textbf{37.50\%}                \\ \hline\bottomrule
\end{tabular}}
\end{center}
\end{table}

\subsection{{Performance on attacks against smoothed classifiers}} \label{sec:attack_against_smoothed_classifiers}

{
We further show the performance on the attack against smoothed classifiers, which is an unseen attack type during training. A smoothed classifier predicts any input $x$ using the majority vote based on   randomly perturbed inputs $\mathcal{N}(x,\sigma^2I)$   \cite{cohen2019certified}. Here we consider the 10-step PGD-$\ell_\infty$ attack generation method with the perturbation radius $\epsilon=20/255$, and $\sigma=0.25$ for smoothing. As shown in Table \ref{table:attack_smoothed_clf}, the proposed CDD-RED method provides the best $\text{PA}_\text{clean}$ and $\text{PA}_\text{adv}$ with a slightly larger $d(\mathbf x, \mathbf x_\mathrm{RED})$ than DO. 
}

\begin{table}[H]
\begin{center}
\caption{{The performance comparison among DO, DS, and {\DeRED} on the PGD attack against smoothed classifiers.}} \label{table:attack_smoothed_clf}
{
\begin{tabular}{c|c|c|c}
\toprule\hline
                                       & DO      & DS      & \DeRED \\ \hline
$d(\mathbf x, \mathbf x_\mathrm{RED})$ & \textbf{15.53}	&22.42&	15.89                   \\ \hline
$\text{\IA}_{\mathrm{benign}}$         & 68.13\%&	70.88\%	& \textbf{76.10\%}               \\ \hline
$\text{\IA}_{\mathrm{adv}}$            & 58.24\%&	58.79\%	& \textbf{61.54\%}               \\ \hline\bottomrule
\end{tabular}}
\end{center}
\end{table}

\section{{Computation cost of RED}} \label{sec:cost}
{We measure the computation cost on a single RTX Titan GPU. The inference time for DO, DS, and {\DeRED} is similar as they use the same denoiser architecture. For the training cost, the maximum training epoch for each method is set as 300. The average GPU time (in seconds) of one epoch for DO, DS, and {\DeRED} is 850, 1180, and 2098, respectively. It is not surprising that {\DeRED} is conducted over a more complex RED objective. Yet, the denoiser only needs to be trained once to reverse-engineer a wide variety of adversarial perturbations, including those unseen attacks during the training.}

\section{{Ablation Studies on {\DeRED}}} \label{sec:ablation}

{In this section, we present additional experiment results using the proposed {\DeRED} method for reverse engineering of deception (RED). We will study the effect of the following model/parameter choice on the performance of {\DeRED}: 1) pretrained classifier $\hat{f}$ for PA regularization, 2) data augmentation strategy  $\check{\mathcal T}$     for PA regularization, and 3) regularization parameter $\lambda$ that strikes a balance between the pixel-level reconstruction error and PA in \eqref{eq: overall_DeRED}.
Recall that the {\DeRED} method  in the main paper sets $\hat{f}$ as VGG19, $\check{\mathcal T} = \{ t \in \mathcal T\,| \,  \hat{F}(t(\mathbf{x})) = \hat{F}(\mathbf{x}),  \hat{F}(t(\mathbf{x}^{\prime})) = \hat{F}(\mathbf{x}^{\prime}) \ \}$, and $\lambda=0.025$. }

\subsection{{Pretrained classifier  \texorpdfstring{$\hat f$}{}.}}

\begin{table}[htb]
\begin{center}
\caption{{The performance of {\DeRED}   using a different   pretrained classifier $\hat{f}$ (either Res50 or R-Res50)  compared with the default setting $\hat{f}=$VGG19.} }\label{table:surrogate_model}
\begin{threeparttable}
{
\begin{tabular}{c|c|c}
\toprule
\hline
 & $\hat{f}$=Res50 & $\hat{f}$=R-Res50\\ \hline
\begin{tabular}[c]{@{}c@{}}$d(\mathbf x, \mathbf x_\mathrm{RED})$\\ (\textcolor{red}{$\downarrow$} is better)\end{tabular} & 12.84 (\textcolor{red}{$\downarrow$ 0.20}) & 10.09 (\textcolor{red}{$\downarrow$ 2.95}) \\ \hline
\begin{tabular}[c]{@{}c@{}}$\text{\IA}_{\mathrm{benign}}$\\ (\textcolor{red}{$\uparrow$} is better)\end{tabular} & 84.33\% ({$\downarrow$1.38\%}) & 57.88\% ($\downarrow$ 27.83\%) \\ \hline
\begin{tabular}[c]{@{}c@{}}$\text{\IA}_{\mathrm{adv}}$\\ (\textcolor{red}{$\uparrow$} is better)\end{tabular} & 79.94\% ($\downarrow$ 0.49\%) & 71.02\% ($\downarrow$ 9.40\%) \\ \hline\bottomrule
\end{tabular}}
\end{threeparttable}
\end{center}
\end{table}

{Besides setting $\hat f$ as VGG19, Table \ref{table:surrogate_model} shows the RED performance using  the other pretrained models, {\it i.e.}, Res50 and R-Res50. As we can see, the use of  Res50 yields the similar performance as VGG19. Although some minor improvements are observed in terms of 
pixel level reconstruction error,  the PA performance suffers a larger degradation.
Compared to  Res50, 
the use of an adversarially robust model 
R-Res50 significantly hampers the RED performance. 
That is because the adversarially robust model typically lowers the prediction accuracy, it is not able to ensure the class-discriminative ability in the non-adversarial context, namely, the {\IA} regularization performance. }

\subsection{{Data selection for {\IA} regularization.}}

\begin{table}[htb]
\begin{center}
\caption{{Ablation study on the selection of $\check{\mathcal T}$ ($\check{\mathcal T}=\mathcal{T}$ and without (w/o) $\check{\mathcal T}$)) for training {\DeRED}, compared with $ \check{\mathcal T} = \{ t \in \mathcal T\,| \,  \hat{F}(t(\mathbf{x})) = \hat{F}(\mathbf{x}),  \hat{F}(t(\mathbf{x}^{\prime})) = \hat{F}(\mathbf{x}^{\prime}). \ \}$}} \label{table:ablation_data}
\begin{threeparttable}
{
\begin{tabular}{c|c|c}
\toprule
\hline
 & $\check{\mathcal T}=\mathcal{T}$ & w/o $\check{\mathcal T}$ \\ \hline
\begin{tabular}[c]{@{}c@{}}$d(\mathbf x, \mathbf x_\mathrm{RED})$\\ (\textcolor{red}{$\downarrow$} is better)\end{tabular} & 15.52 ($\uparrow$ 2.48)  & 13.50 ($\uparrow$ 0.46)  \\ \hline
\begin{tabular}[c]{@{}c@{}}$\text{\IA}_{\mathrm{benign}}$\\ (\textcolor{red}{$\uparrow$} is better)\end{tabular} & 83.64\% ($\downarrow$ 2.07\%) &  84.04\% ({$\downarrow$1.67\%}) \\ \hline
\begin{tabular}[c]{@{}c@{}}$\text{\IA}_{\mathrm{adv}}$\\ (\textcolor{red}{$\uparrow$} is better)\end{tabular} & 75.92\% ($\downarrow$ 4.51\%) & 79.99\% ($\downarrow$ 0.44\%)  \\ \hline\bottomrule
\end{tabular}}
\end{threeparttable}
\end{center}
\end{table}

{As data augmentation might alter the classifier's original decision in \eqref{eq: overall_DeRED},  we study how $\check{\mathcal T}$ affects the RED performance by setting   $\check{\mathcal T}$ as the original data, i.e., without data augmentation, and all data, i.e., $\check{\mathcal T} = \mathcal{T}$. Table \ref{table:ablation_data} shows the performance of different $\check{\mathcal T}$ configurations, compared with the default setting. 
The performance is measured on the testing dataset. As we can see, using all data or original data cannot provide an overall better performance than {\DeRED}. That is because the former might cause over-transformation, and the latter lacks generalization ability. }

\subsection{{Regularization parameter  \texorpdfstring{$\lambda$}{}.}}

\begin{table}[htb]
\begin{center}
\caption{{Ablation study on the regularization parameter $\lambda$ (0, 0.0125, and 0.05) for {\DeRED} training, compared with $\lambda$=0.025. }} \label{table:ablation_lambda}
\begin{threeparttable}
{
\begin{tabular}{c|c|c|c}
\toprule\hline
 & $\lambda$=0 & $\lambda$=0.0125 & $\lambda$=0.05 \\ \hline
\begin{tabular}[c]{@{}c@{}}$d(\mathbf x, \mathbf x_\mathrm{RED})$\\ (\textcolor{red}{$\downarrow$} is better)\end{tabular} & 8.92 (\textcolor{red}{$\downarrow$ 4.12}) & 12.79 (\textcolor{red}{$\downarrow$ 0.25}) & 14.85 ($\uparrow$ 2.13) \\ \hline
\begin{tabular}[c]{@{}c@{}}$\text{\IA}_{\mathrm{benign}}$\\ (\textcolor{red}{$\uparrow$} is better)\end{tabular} & 47.61\% ({$\downarrow$38.10\%}) & 81.00\% ({$\downarrow$4.71\%}) & 85.56\% ({$\downarrow$ 0.15\%)} \\ \hline
\begin{tabular}[c]{@{}c@{}}$\text{\IA}_{\mathrm{adv}}$\\ (\textcolor{red}{$\uparrow$} is better)\end{tabular} & 73.37\% ({$\downarrow$7.06\%}) & 78.25\% ($\downarrow$ 2.18\%) & 79.94\% ($\downarrow$ 0.49\%) \\ \hline\bottomrule
\end{tabular}}
\end{threeparttable}
\end{center}
\end{table}

{The overall training objective of {\DeRED} is (\ref{eq: overall_DeRED}), which is the weighted sum of the reconstruction error and PA  with a regularization parameter $\lambda$. We further study the sensitivity of {\DeRED} to the choice of 
$\lambda$, which is set by  $0, 0.0125$, and $0.05$. Table \ref{table:ablation_lambda} shows the RED performance of using different $\lambda$ values, compared with the default setting $\lambda=0.025$. We report the average performance on the testing dataset. As we can see, the use of $\lambda=0$, which corresponds to training the denoiser without PA regularization, achieves a lower pixel-level reconstruction error $d(\mathbf x, \mathbf x_\mathrm{RED})$, but   degrades the prediction-level performance, especially $\text{\IA}_{\mathrm{benign}}$ greatly. In the same time, $\lambda=0$ provides a smaller pixel-level reconstruction error with a better PA performance than DO, which indicates the importance of using proper augmentations. We also observe that keep increasing  $\lambda$ to $0.05$ is not able to provide a better PA.} 

\section{{Ablation study of different attack hyperparameter settings}}
\label{sec:diff_attack_hyper}

{We test on PGD attacks generated with different step sizes, including $4/255$ and $6/255$, and with and without random initialization (RI). Other hyperparameters are kept the same. The adversarial examples are generated by the same set of images w.r.t. the same classifier ResNet-50. The results are shown in Table \ref{table:diff_pgd_hyper}. As we can see, the RED performance is quite robust against the varying hyperparameters of PGD attacks. Compared with DO, CDD-RED greatly improves $\text{PA}_{\text{benign}}$ and achieves higher $\text{PA}_{\text{adv}}$ with a slightly larger $d(\mathbf x, \mathbf x_\mathrm{RED})$. Compared to DS, {\DeRED} achieves slightly better PA but with a much smaller reconstruction error $d(\mathbf x, \mathbf x_\mathrm{RED})$.} 
\begin{table}[H]
    \centering
        \caption{{RED performance for PGD with different hyperparameter settings, including stepsize as $4/255$ and $6/255$, and with and without RI.}}
    \label{table:diff_pgd_hyper}
    \begin{tabular}{cc}
 \begin{tabular}{c|c|c|c|c}
\toprule\hline
Without RI / With RI                                     & Stepsize & DO                                                             & DS                                                             & CDD-RED                                                        \\ \hline
                                                         & 4/255    & \textbf{5.94/5.96}                                                      & 16.56/16.57                                                    & 8.91/8.94                                                      \\ \cline{2-5} 
\multirow{-2}{*}{$d(\mathbf x, \mathbf x_\mathrm{RED})$} & 6/255    & \textbf{5.99/5.98}                                                      & 16.52/16.50                                                    & 8.97/8.94                                                      \\ \hline
                                                         & 4/255    & 40.00\%/47.00\%                                                & 91.00\%/91.00\%                                                &\textbf{ 94.00\%/93.00\%  }                                              \\ \cline{2-5} 
\multirow{-2}{*}{$\text{\IA}_{\mathrm{benign}}$}         & 6/255    & 51.00\%/61.00\%                                                & 94.00\%/93.00\%                                                & \textbf{95.00\%/94.50\%}                                                \\ \hline
                                                         & 4/255    &  97.50\%/97.50\% &  97.50\%/97.50\% & \textbf{ 99.50\%/99.50\%} \\ \cline{2-5} 
\multirow{-2}{*}{$\text{\IA}_{\mathrm{adv}}$}            & 6/255    & 96.50\%/96.50\% &  95.50\%/95.50\% & \textbf{98.50\%/99.50\%} \\ \hline\bottomrule
\end{tabular}
    \end{tabular}
\end{table}

\section{{Performance without adversarial detection assumption}} \label{sec:mixture}

 {The focus of RED is to demonstrate the feasibility of recovering the adversarial perturbations from an adversarial example. However, in order to show the RED performance  on the global setting, we experiment with a mixture of 1) adversarial images, 2) images with Gaussian noise (images with random perturbations), and 3) clean images on the ImageNet dataset. The standard deviation of the Gaussian noise is set as 0.05. Each type of data accounts for \textbf{1/3} of the total data. The images are shuffled to mimic the live data case. The overall accuracy before denoising is 63.08\%. After denoising, the overall accuracy obtained by DO, DS, and CDD-RED is 72.45\%, 88.26\%, and \textbf{89.11\%}, respectively. During the training of the denoisers, random noise is not added to the input. }
 
 


\section{{Transferability of reconstructed adversarial estimate}} \label{sec:tranferability}

{We further examine the transferability of RED-synthesized perturbations. In the experiment, RED-synthesized perturbations are generated from PGD attacks using ResNet-50. We then evaluate the attack success rate (ASR) of the resulting perturbations transferred to the victim models ResNet-18, VGG16, VGG19, and Inception-V3. The results are shown in Table \ref{table:transferability}. As we can see, the perturbation estimates obtained via our {\DeRED} yield better attack transferability than DO and DS. Therefore, such RED-synthesized perturbations can be regarded as an efficient transfer attack method.}

\begin{table}[H]
\begin{center}
\caption{{The tranferability of RED-synthesized perturbations.}} \label{table:transferability}
{
\begin{tabular}{c|c|c|c}
\toprule\hline
                                       & DO      & DS      & \DeRED \\ \hline
ResNet-18 & 66.50\%	& 70.50\% &	 \textbf{77.50\%}               \\ \hline
VGG16        & 71.50\%	& 74.00\% &	\textbf{81.00\%}              \\ \hline
VGG19        & 71.50\%	& 70.00\%	& \textbf{80.00\%}               \\ \hline
Inception-V3            & 86.00\% &	85.50\%	& \textbf{ 90.00\%}               \\ \hline\bottomrule
\end{tabular}}
\end{center}
\end{table}

\section{{potential applications of RED}} \label{section:potential_app}

{In this paper, we focus on recovering attack perturbation details from adversarial examples. But in the same time, the proposed RED can be leveraged for various potential interesting  applications. In this section, we delve into three applications, including RED for adversarial detection in \ref{sec:detection}, inferring the attack identity in \ref{sec:identity}, and provable defense in \ref{sec:provable}. }

\subsection{{RED for adversarial detection }} \label{sec:detection}
\begin{figure}[H]
    \centering
    \includegraphics[width=0.5 \columnwidth]{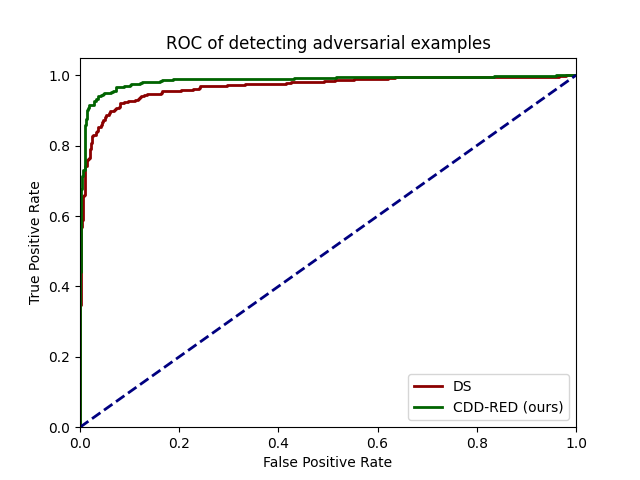}
    \caption{{RoC of detecting adversarial examples.}}
    \label{fig:RoC}
\end{figure}

 {The outputs of RED can be looped back to help the design of adversarial detectors. Recall that our proposed RED method (CDD-RED) can deliver an attribution alignment test, which reflects the sensitivity of input attribution scores to the pre-RED and post-RED operations. Thus, if an input is an adversarial example, then it will cause a high attribution dissimilarity (i.e., misalignment) between the pre-RED input and the post-RED input, i.e., $I(x, f(x))$ vs. $I(D(x), f(D(x)))$ following the notations in Section \ref{sec:evaluation_metric}. In this sense, attribution alignment built upon $I(x, f(x))$ and $I(D(x), f(D(x)))$ can be used as an adversarial detector. Along this direction, we conducted some preliminary results on RED-assisted adversarial detection, and compared the ROC performance of the detector using CDD-RED and that using denoised smoothing (DS).  In Figure \ref{fig:RoC}, we observe that the CDD-RED based detector yields a superior detection performance, justified by its large area under the curve. Here the detection evaluation dataset is consistent with the test dataset in the evaluation section of the paper.}
 
\subsection{{RED for attack identity inference}}\label{sec:identity}

\begin{figure}[H]
    \centering
    \includegraphics[width=1 \columnwidth]{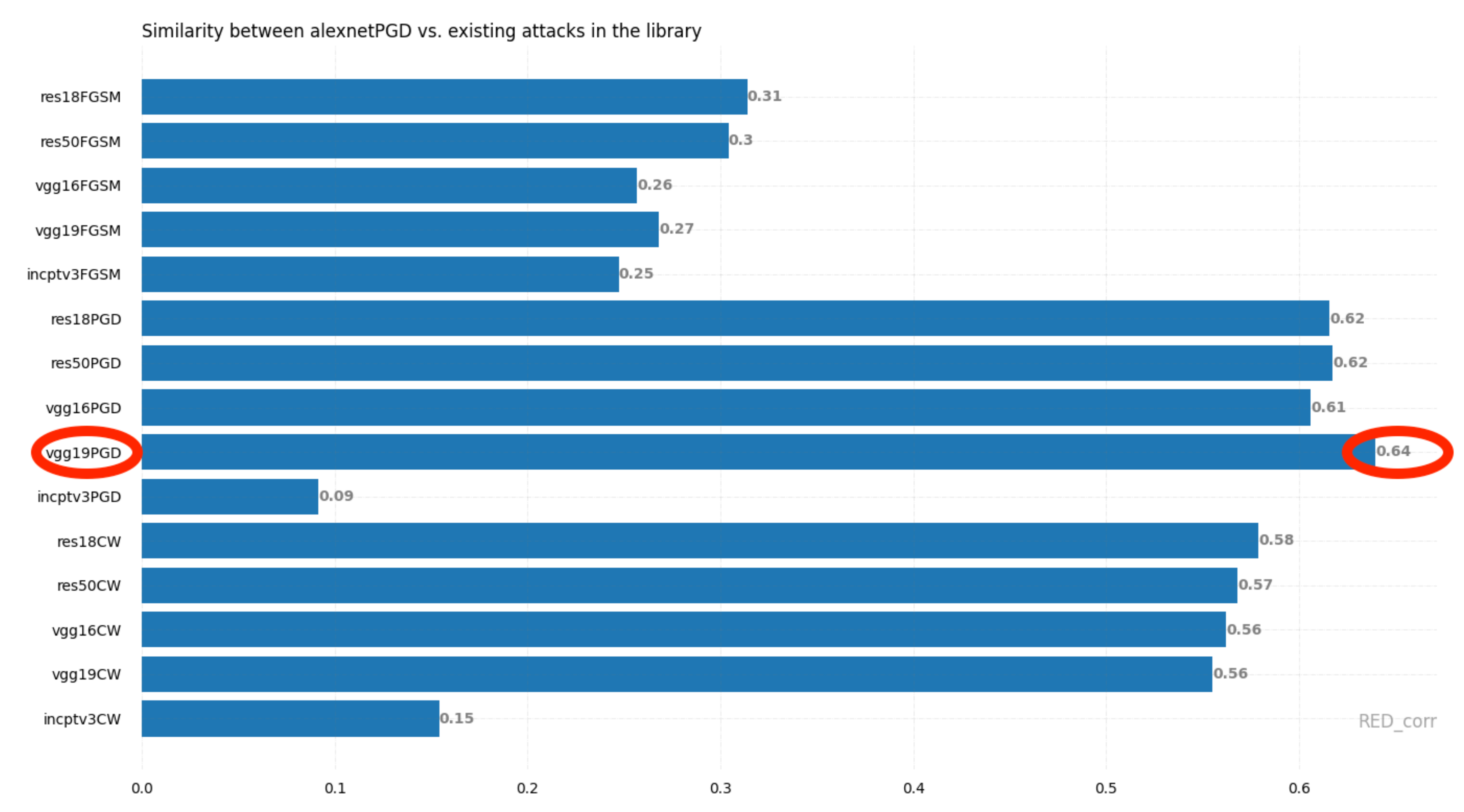}
    \caption{{Similarity between AlexNet-generated PGD vs. existing attacks in the library.}}
    \label{fig:identity}
\end{figure}

{We consider another application to infer the attack identity using the reverse-engineered adversarial perturbations. Similar to the setup of Figure \ref{fig:correlationmatrix}, we achieve the above goal using the correlation screening between the new attack and the existing attack type library. Let $z^\prime$  (e.g., PGD attack generated under the unseen AlexNet victim model) be the new attack. We can then adopt  the RED model $D(\cdot)$ to estimate the perturbations $\delta_{new} = z^\prime - D(z^\prime)$. And let $x^\prime_{Atk_i}$ denote the generated attack over the estimated benign data $D(z^\prime)$ but using the existing attack type i. Similarly, we can obtain the RED-generated perturbations $\delta_{i} = x^\prime_{Atk_i} - D(x^\prime_{Atk_i})$. With the aid of $\delta_{new}$ and $\delta_{i}$ for all $i$, we infer the most similar attack type by maximizing the cosine similarity: $i^* = \text{argmax}_{i} ~ cos(\delta_{new},\delta_{i})$. Figure \ref{fig:identity} shows an example to link the new AlexNet-generated PGD attack with the existing VGG19-generated PGD attack. The reason is elaborated on below. (1) Both attacks are drawn from the PGD attack family. And (2) in the existing victim model library (including ResNet, VGG, and InceptionV3), VGG19 has the most similar architecture as AlexNet, both of which share a pipeline composed of convolutional layers following fully connected layers without residual connections.}

\subsection{{RED for provable defense}} \label{sec:provable}

\begin{figure}[H]
    \centering
    \includegraphics[width=0.6\columnwidth]{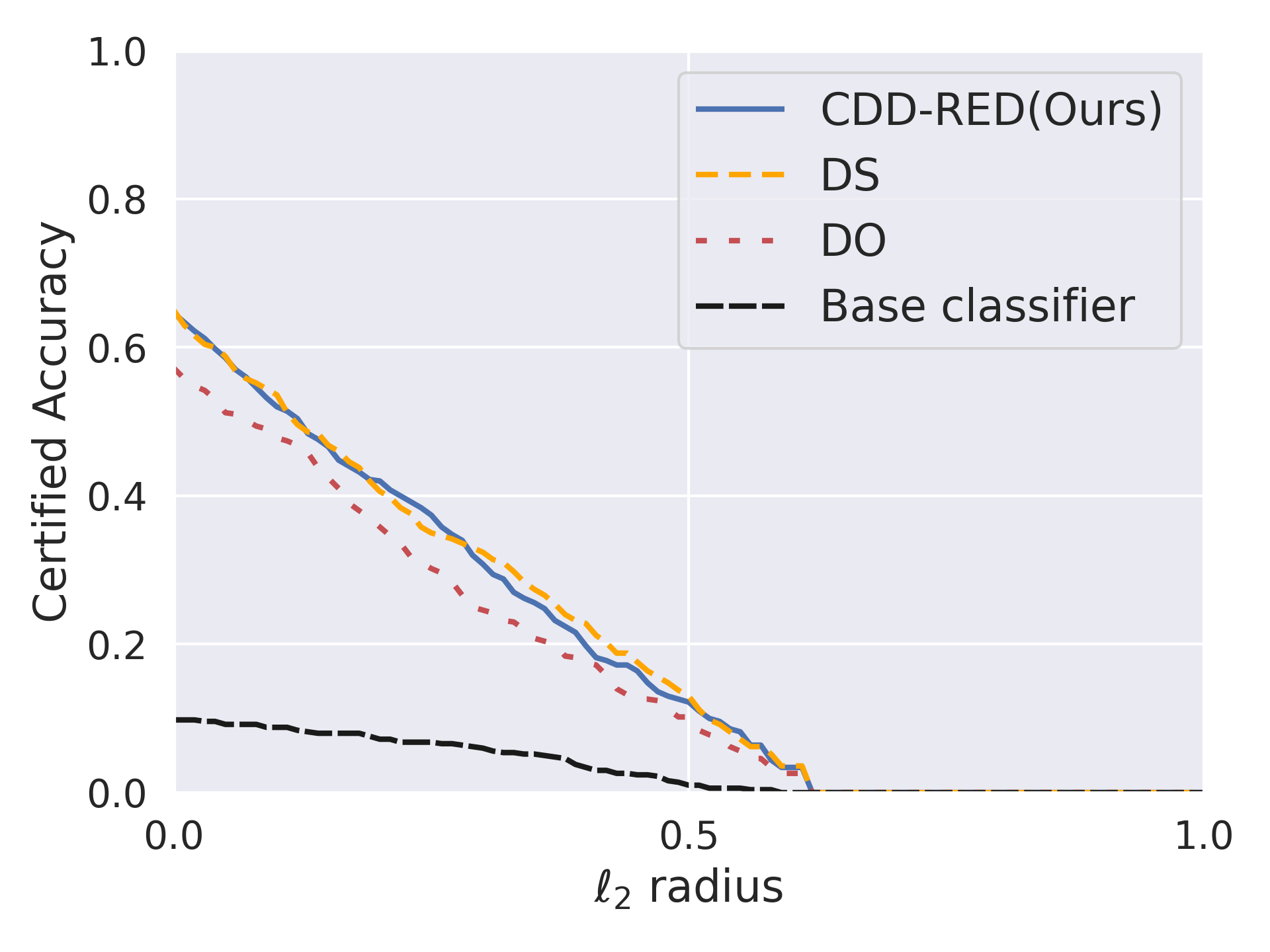}
    \caption{{Certified Robustness by different methods.}}
    \label{fig:certified}
\end{figure}

{We train the RED models to construct smooth classifiers, the resulting certified accuracy is shown in Figure \ref{fig:certified}. Here the certified accuracy is defined by the ratio of correctly-predicted images whose certified perturbation radius is less than the $\ell_2$ perturbation radius shown in the x-axis.}

\end{document}